\documentclass[journal]{IEEEtran}
\usepackage{cite}
\usepackage{pdflscape}
\usepackage{afterpage}
\usepackage{capt-of}
\usepackage{amsmath,amssymb,amsfonts}
\usepackage{algorithmic}
\usepackage{rotating}
\usepackage{graphicx}
\usepackage{textcomp}
\usepackage{lipsum}
\usepackage{authblk}
\usepackage[table,xcdraw]{xcolor}
\usepackage{url}
\usepackage{multirow}
\usepackage[flushleft]{threeparttable}
\begin{document}

%\title{Leveraging Transfer Learning and Data Augmentation in a Multi-Scale Segmentation Network to Detect Skin Lesion and its Attributes in Dermoscopy Images}
\title{Segmentation of Skin Lesions and their Attributes Using Multi-Scale Convolutional Neural Networks  and Domain Specific Augmentations}
%\title{Leveraging Multi-Scale Convolutional Neural Networks  and Domain Specific Augmentations for Skin Lesions and Attributes Segmentation}

\author{Mostafa~Jahanifar*, %~\IEEEmembership{Student Member,~IEEE,}
        Neda~Zamani~Tajeddin*, %~\IEEEmembership{Student Member,~IEEE,}
        Navid~Alemi~Koohbanani*, %~\IEEEmembership{Student Member,~IEEE,}
        Ali~Gooya, and  %~\IEEEmembership{Member,~IEEE,}
        Nasir~Rajpoot%~\IEEEmembership{Senior Member,~IEEE}% <-this % stops a space
\thanks{M. Jahanifar and N. Zamani Tajeddin were with the Department
of Biomedical Engineering, Tarbiat Modares University. M. Jahanifar is now with the Department of Research and Development, NRP company, Tehran, Iran (e-mails: \{m.jahanifar,n.zamanitajeddin\}@modares.ac.ir).}% <-this % stops a space
\thanks{N. Alemi Koohbanani and N. Rajpoot are with the Tissue Image Analytics (TIA) Lab at the Department of Computer Science, University of Warwick, UK (emails: \{n.alemi-koohbanani,N.M.Rajpoot\}@warwick.ac.uk).}% <-this % stops a space
\thanks{Ali Gooya is with the Department of Electronic and Electrical Engineering, University of Sheffield, Sheffield, England (e-mail: a.gooya@sheffield.ac.uk)}
\thanks{*These authors contributed equally to this work}}

% The paper headers
\markboth{}
{Shell \MakeLowercase{\textit{Jahanifar et al.}}: Leveraging Transfer Learning and Data Augmentation in a Multi-Scale Segmentation Network to Detect Skin Lesion and Dermoscopic Attributes}

% make the title area
\maketitle

\begin{abstract}
Computer-aided diagnosis systems for classification of different type of skin lesions have been an active field of research in recent decades. %Recent approaches utilize convolutional neural networks (CNNs) to do such a task.
It has been shown that introducing lesions and their attributes’ masks into lesion classification pipeline can greatly improve the performance. In this paper, we  propose a framework  by incorporating transfer learning  for segmenting  lesions and their attributes  based on the convolutional neural networks. The proposed framework is based on the  encoder-decoder architecture which utilizes a variety of pre-trained networks in the encoding path and generates the prediction map by combining multi-scale information in decoding path using a pyramid pooling manner. To address the lack of training data and increase the proposed model generalization, an extensive set of novel domain-specific augmentation routines have been applied to simulate the real variations in dermoscopy images. Finally, by performing broad experiments on three different data sets obtained from International Skin Imaging Collaboration archive (ISIC2016, ISIC2017, and ISIC2018 challenges data sets), we show that the proposed method outperforms other state-of-the-art approaches for ISIC2016 and ISIC2017 segmentation task and achieved the 1\textsuperscript{st} rank on the leader-board of ISIC2018 attribute detection task.
\end{abstract}


\section{Introduction}
\IEEEPARstart{S}{kin} lesion analysis through image processing and machine learning techniques has attracted a lot of attention in the last decades \cite{oliveira2016computational}. With the advent of deep convolutional neural networks (CNNs) and different scientific competitions on this topic in the recent years \cite{gutman2016skin,codella2017skin}, numerous powerful computational methods have been emerged to solve different problems in this area. In general, the final goal of using computer aided diagnosis (CAD) systems for skin lesion analysis is to automatically detect any abnormalities or skin diseases. The methods in the literature usually follow a certain analysis pipeline \cite{tajeddin2018melanoma}: first, lesion area in the image is detected (segmentation task) which secondly helps other computerized algorithms to extract discriminative feature (feature extraction task) from the lesion,  to finally decide about the type of skin lesions (classification task). %Therefore, one of the main steps in lesion analysis algorithms is to segment the lesion boundary from the normal skin.
In this framework, precise segmentation of lesion boundary is a critical task which can lead to representative features to differentiate between normal and malignant lesions.
 
%which leads to extraction of relevant features, otherwise, extracted features may represent the normal skin as well  and confuse the lesion classification (decease recognition) task \cite{yu2017TMI}.

Lesions attributes, also known as dermoscopic structures, are meaningful visual patterns in the skin lesion texture. Each of them represents a specific diagnostic symptom, and based on the lesion type and malignancy severity their appearance may change \cite{braun2019attributes}. Some lesions may not contain any specific dermoscopic pattern  while some may include textural pattern related to several attributes. There are different types of attributes defined in dermatology. Five of the most clinically interesting attributes are pigment networks, globules/dot, Milia like cysts, negative networks, and streaks. In clinical practices, apart from the general and local appearance characteristics of the lesion, dermatologists look for these attributes to better decide about the lesion type and its malignancy level. Hence, CAD systems can also benefit from the useful information that attributes deliver, to achieve a better automatic recognition performance.

Although modern CNNs are able to extract relevant features from the raw dermoscopic images and classify them without the need of lesions or their attributes segmentation map \cite{litjens2017survey}, it has been shown that incorporating additional information can significantly improve lesion classification task \cite{yu2017TMI,diaz2018dermaknet}. Yu et al. \cite{yu2017TMI} showed that incorporating lesion mask can elevate lesion classification and they achieved 1\textsuperscript{st}  rank in the ISIC2016: Melanoma Recognition challenge. Moreover, Gonzalez Diaz \cite{diaz2018dermaknet} proposed a lesion classification framework that incorporates both lesion segmentation and attributes probability maps which was able to achieve state-of-the-art results using only one single model by adding this additional information. Such researches and their promising results prove the importance of lesion mask and its attributes segmentation tasks for diagnosis purposes.

There are various hurdles in automatic analysis of dermoscopic images due to the artefact in image capturing setup, sample preparations and lesion appearance itself. During image capturing from skin lesion samples using a dermatoscope  different image artifact can affect the output quality. These artefacts are hair occlusion, color charts, ruler markers, low image contrast and sharpness, uneven illumination, gel bubbles, marker annotations, clothing obstacles, color inconstancy, and lens artifact (dark corners). Besides, skin lesions can show large variability in appearance based on their type. Severe cases may have fuzzy borders, complex or vague texture, low contrast against the skin, amorphous geometry, and multi-colored face. Therefore, designing a general algorithm to overcome all these hurdles is hardly achievable. Attribute segmentation task is even more challenging due to their small structures and weak appearance.

Here, we are looking for powerful CNN models to segment lesion and its related attributes from dermoscopic images. In order to avoid the complications during training a CNN from scratch, we propose to use transfer learning in a novel segmentation framework. Our proposed segmentation network benefits from multi-scale convolutional and pyramid pooling blocks to detect desired small and big objects from the image.
Another novel aspect of the present work is extensive use of augmentation routines. Considering the domain-specific knowledge, we design some exquisite augmentations techniques to imitate the appearance variation in dermoscopy images. Prediction maps are then generated by ensembling the outputs of different models and different test time augmentation. To obtain the binary mask of the lesion or dermoscopy attribute, optimal thresholding and post-processing tasks are applied on the final ensemble prediction map.

Rest of the paper is organized as follows: 
In section \ref{sec:literature}, we broadly cover the recent literature for segmenting skin lesion and its attributes which is followed by description of our approach to tackle these problems in section \ref{sec:Methodology}. Section \ref{sec:setups} comprises some explanation about the data set, evaluation metrics, and implementation details. Experimental results for lesion segmentation and attributes detection tasks on three different benchmark data sets are reported in  section \ref{sec:setups}. Finally, the  discussion and conclusion of this research are drawn in sections \ref{sec:discussion} and  \ref{sec:conclusion}, respectively.

%%%%%%%%%%%%%%%%%%%%%%%%%%%%%%%%%%%%%%%%%%%%%%%%%%%%%%%%%%%%%%%%%%%%%%%%%%%%%%%%%%%%%%%%%
\section{Literature Review} \label{sec:literature}

In the following, we extensively review the latest researches for skin lesion and their attributes segmentation.
There are several methods introduced in the literature to address lesion segmentation and it has been an active field of research for several decades. Some are based on thresholding algorithms, some followed region based methods like clustering algorithms, and hybrid approaches that suggest refining an initial segmentation through another algorithms (like active contours, region growing, morphological operations, etc.).  However, due to the prominent result of deep learning, majarity of recent successful approaches for skin lesion segmentation are based on deep convolutional neural networks (CNNs) and the well-known network architectures in computer vision. %\cite{yuan2017TMI,yu2017TMI,diaz2018dermaknet,esteva2017dermatologist}.
In the current literature review, we are focusing on the methods that reported their performance at least on one of the ISIC2016, ISIC2017, or ISIC2018 data sets \cite{gutman2016skin,codella2017skin,Tschandl2018_HAM10000}, in order to put together benchmark tables for those data sets. To have a more comprehensive review of different lesion segmentation algorithms, enthusiastic readers can refer to \cite{oliveira2016computational, Celebi2015}.

\subsection{Lesion segmentation}
%%%%%%%% starts  saliency
\subsubsection{saliency detection based methods}
Dealing with lesion segmentation as a saliency detection problem has been exploited several times \cite{jahanifar2018supervised,fan2017,ahn2017}. In our previous work \cite{jahanifar2018supervised}, we proposed to construct the saliency map of the lesion in a supervised and multi-scale manner based on the shape, color, and textures features extracted from super-pixel segments of the images. In \cite{jahanifar2018supervised} several regression models in different scales predict initial saliency maps based on extracted features and the final saliency map of the lesion is achieved through combining them. 
In a recent paper, Garcia \textit{et al.} \cite{jose2019} proposed a hybrid segmentation method with two main parts: pixel fuzzy classification and histogram thresholding. Their method can be consider as a saliency detection approach like \cite{jahanifar2018supervised} since authors in \cite{jose2019} extracted some pixel-wise color and texture features from image and allocate each pixel to one of lesion, skin, or other probability maps (saliency maps). Afterward, they employed a novel histogram-based thresholding algorithm on the probability maps to achieve the lesion segmentation.

Ahn \textit{et al.} \cite{ahn2017} also took a supervised saliency detection approach. They inverted the lesion detection into a background detection problem. In their method, the background map is reconstructed by analysing image's super-pixels in a multi-scale framework. Then, they proposed that lesion area can be drawn from super-pixels having the highest reconstruction error. They pre-processed the input images to remove hairs and color chart from the images and incorporated thresholding and post-processing operations on the output \cite{ahn2017}.  
In an unsupervised fashion, Fan \textit{et al.} \cite{fan2017} introduced two saliency detection routines based on color and brightness information and proposed a method to fuse these saliency maps in order to construct the final lesion prediction. Then, the segmentation process is done through a histogram-based thresholding algorithm.

%%%%%%%% starts  unsupervised
\subsubsection{Unsupervised hybrid methods}
Unsupervised hybrid methods that usually contain several steps and integrate different algorithms have been investigated in the literature \cite{tajeddin2016general, guo2018, navarro2018}.
Zamani \textit{et al.} \cite{tajeddin2016general, tajeddin2018melanoma} proposed a thresholding algorithm accompanied by multiple advanced pre-processing and post-processing steps to ahcieve an initial prediction of the lesion boundary. They then proposed to use a dual-component speed function, based on the image intensity and color information to drive an active contour towards the real boundary of the lesion. 
Guo \textit{et al.} \cite{guo2018} proposed a hybrid segmentation algorithm which comprises three main stages: First,  Shearlet transform is applied on the red channel of the input image  to construct three sets (channels) of brightest, intermediate, and non-white pixels. Second, an indeterminacy filter is applied on the intermediate channel values and a Neutrosophic C-Means clustering algorithm is then employed to detect the initial mask of the lesion. Third, the initial mask is evolved  through an adaptive region growing algorithm to construct the final segmentation of the lesion. 
Another unsupervised hybrid approach is proposed in \cite{navarro2018} which first segments the image into super-pixels using local features guided SLIC algorithm and then iteratively merge those regions to form two classes of regions (lesion and non-lesion), considering a spatial continuity constraint on the super-pixels' color.

%%%%%%%% Deep learning based method
\subsubsection{Convolutional neural network based methods}
Based on our review, apart from the above-mentioned articles, rest of the literature used deep learning based approaches for lesion segmentation (applied on ISCI2016,2017 and 2018). It seems that most segmentation networks are designed in an encoder-decoder paradigm. Based on the network architecture, we reviewed these methodd in three categories: 1) methods that are based on well-known FCN architecture \cite{FCN} where usually the final feature maps in the encoding path are upsampled to the size of the input image in one or two levels and no many features maps are extracted in the decoding path. These architectures usually incorporate one or two skip connections. 2) network architectures that are like U-Net model \cite{UNET} which accomplish feature extraction and segmentation through consecutive convolutional layers in several levels of  the encoding and decoding paths and usually conjoin the feature maps of encoding path to the feature maps of the corresponding levels on the decoding path using skip connections. 3) The third group of architectures are convolutional-deconvolutional models that can be considered as special cases of U-Net architectures, with transposed convolution layers in their decoding path instead of convolutional layers. 

Galdran \textit{et al.} \cite{galdran2017} used the original U-Net model \cite{UNET} for lesion segmentation, however the novelty in their work was related to their proposed domain-specific augmentation techniques. They used shades of gray color constancy techniques to normalize the color throughout the data set images while retaining the estimated illuminants information, enabling them to randomly change the color and illumination of normalized images during the training phase.
Berseth \cite {berseth2017} used a U-shaped neural network that, unlike the original U-Net architecture \cite{UNET}, is not fully convolutional i.e., his proposed network includes three down-sampling and up-sampling levels which are connected by a fully connected layer. Xu \textit{et al.} \cite{xu2018} also employed an encoder-decoder architecture benefiting form their proposed hybrid convolution, down-sampling, and up-sampling building blocks. The general paradigm of their network is like U-Net architecture which has consecutive implementation of their proposed building blocks in each level of encoding and decoding paths, however, they didn't implement any skip connection in their architecture.

Several models have empowered their segmentation models using transfer learning concept and incorporating pre-trained models into their network architecture. In a recent work, Tschandl \textit{et al.} \cite{tschandl2019} proposed a U-Net style network in which a ResNet34 architecture \cite{ResNet}, that was previously fine-tuned on a lesion classification task, has been embedded in the encoding path. Moreover, a new decoder block has been introduced for the decoding path in \cite{tschandl2019}, which uses two convolution layers and a transposed convolution layer between them.
Sheng Chen \textit{et al.} \cite{chen2018} claimed that solving skin lesion classification and segmentation tasks synergistically can improve the performance. Therefore, they proposed a two-branched network in which a  ResNet101 model \cite{ResNet} is used to extract features from the input image and those features are mutually fed into segmentation and classification branches. Moreover, authors in \cite{chen2018} introduced a feature passing module that  selects useful feature maps in classification branch and passes them to the segmentation branch and vice versa. Mengola \textit{et al.} \cite{menegola2017} embedded a pre-trained VGG16 model \cite{VGG} in the encoding path of their  UNet-like segmentation architecture. However, unlike the original U-Net \cite{UNET} which is fully convolutional, they kept the fully connected layer of the VGG model in the encoding path similar to \cite{xu2018}.

 In a network proposed by Sarker \textit{et al.} \cite{sarker2018}, the encoder part is built of four pre-trained dilated residual networks and a pyramid pooling block which enables the network to capture multi-scale feature maps and the decoder path consists of convolutional and upsampling layers. They also introduced a novel segmentation loss based on softmax and incorporated only one skip connection between the encoding and decoding paths \cite{sarker2018}. 
 Mirikharaji et al. \cite{mirikharaji2018} claimed that using a posteriori information besides the original appearance information can elevate and refine the segmentation results. Therefore, they used multiple U-Net models \cite{UNET} in a consecutive fashion, in which the input for each U-Net model is the original image concatenated with the output prediction of the previous U-Net model.
 In an adversarial learning framework, Xue \textit{et al.} \cite{xue2018} proposed an end-to-end approach for segmentation of skin lesion. Their method utilized two different neural networks: Segmentor network to detect skin lesions from images and critic network that measures the difference of output segmentation with ground truth. Critic network duty is to force the Segmentor network to perform better and create more precise results. Training of the whole architecture is done in an adversarial manner, therefore, the segmentor network weights are updated based on the gradients computed from the Critic network's loss function. Authors employed a fully convolutional encoder-decoder architecture for segmentor and another simple convolutional one for the critic. Therewith, they incorporated a new multi-scale $L_1$ loss function for adversarial training procedure. 
 
 Xiaomeng Li \textit{et al.} \cite{xiaomeng2018} proposed a semi-supervised framework in which a segmentation network is optimized for a supervised loss function (using labeled data) and an unsupervised loss (calculated based on the prediction from unlabeled data) as a regularisation term. The unsupervised training part procedure is done through a consistent transformation scheme where the prediction resulted from the original input image is transformed (e.g. rotated) and compared with a transformed version of the ground truth (through a cross-entropy loss) and in another try, original input is firstly transformed and fed into the network, then its  prediction is compared with the last rotated prediction resulted from the original labeled image (through a mean square error). The final loss is constructed based on the weighted compilation of the supervised and unsupervised losses. Xiaomeng et al. chose a dense version of UNet as their segmentation model.
Al-masani \textit{et al.} \cite{almasani2018} proposed a full resolution convolution neural network without any pooling or upsampling layers. Their model extracts full resolution feature maps in several convolutional layers.
%%%% End of UNET like

Many articles used FCN-like  architectures as their segmentation model like Kawahara \textit{et al.} \cite{kawahara2018fully} in which the base network is VGG16 \cite{VGG} pre-trained on ImageNet data set and its fully connected layers are removed. 
Yuexiang Li et al. \cite{li2018} addressed the problem of lesion segmentation and classification, simultaneously. Their proposed framework is constructed of two fully convolutional residual networks (FCRN) that employs a modified version of residual blocks and takes input images in two different resolutions. The final segmentation is constructed through the summation of interpolated outputs from each FRCN and a lesion index calculation unit \cite{li2018}.
A two-part detection-segmentation learning framework was introduced by Qian \textit{et al.} \cite{qian2018}. They proposed to first detect the most probable lesion region from the image using Mask R-CNN model \cite{he2017mask}, next crop that region from the image and resize it to a fixed certain resolution (512 $\times$ 512), finally that patch is fed to a FCN-like segmentation network to achieve fine boundary of the lesion. Their proposed segmentation network is equipped with an atrous spatial pyramid pooling (ASPP) block at the end of encoding path to achieve high resolution output.

Lei Bi \textit{et al.} used FCN architecture in different ways in their papers. In their first work, a semi-automatic approach is proposed using a multi-scale manner. The user interaction (specialist) is applied by manually clicking on two points: one inside and another outside of the lesion area. Based on the clicked points, two Euclidean distance maps are created that are fed into the multi-scale FCN framework alongside the original image.
In another work  \cite{bi2017TBME}, they used FCN architecture  several times in a multi-stage framework to better segment the lesion boundary. The first stage learns coarse prediction and rough localization of the lesion then, in the next stage, more details are predicted by leveraging the knowledge acquired from the previous stages by using a parallel integration mechanism to include the information from all primary stages.
In their next article, Bi \textit{et al.}  \cite{bi2017ISIC} employed transfer learning concept in their proposed architecture. First , a pre-trained ResNet \cite{ResNet} model is fine-tuned on skin lesion images and afterwards upsampling layers were added to form an FCN-like architecture for lesion segmentation goal. Comparing to other approaches applied on ISIC2017 data set they had the benefit of  using 7800 more training images from external sources.
However, in their most recent work Bi \textit{et al.} \cite{bi2019} proposed a class-specific segmentation method where they trained three fully convolutional segmentation network based on their previous work \cite{bi2017ISIC}: two networks were trained separately for melanoma and non-melanoma images and the third one was trained on a set of both classes. Afterwards, they used a probability base step-wise integration module to combine the outcomes from those three segmentation networks. The probability for mixing the predictions comes from a basic lesion classification network based on the ResNet \cite{ResNet} architecture.

The second method \cite{yu2017TMI} on ISIC2016 Task1 leader-board was also based on a fully convolutional residual network in a FCN-like architecture. Segmentation network in \cite{yu2017TMI} implements a model with residual blocks in the encoder and strided deconvolutions in the decoder path. Their network has two skip connections with bottleneck convolutions. 

%%%% End of FCN like

Some researches introduced convolutional-deconvolutional architectures in an  encoder-decoder paradigm to segment skin lesions. As the name indicates, these models have mostly transposed convolution (deconvolution) layers in their decoding path. 
Yading Yuan \textit{et al.} \cite{yuan2017TMI, yuan2017} proposed to use this type of network architecture in their recent researches using  kernels with different sizes and dropout in both encoding and decoding paths. However, the main contribution of their work was introducing a novel loss function inspired by Jaccard distance, which was able to produce smoother and more accurate predictions. They were able to rank first in the ISIC2017 Task1 Lesion Segmentation challenge using their most recent method \cite{yuan2017}.

In the same convolutional-deconvolutional paradighm, Hang Li \textit{et al.} \cite{li2018JBHI} utilized pre-trained ResNet model \cite{ResNet} to extract abstract multi-scale features from the input image and then construct the segmentation map through the decoding path. Dense deconvolutional layers, chained residual poolings and auxiliary losses in two intermediate levels where utilized in the decoding path.
He \textit{et al.} \cite{he2018} proposed a multi-path deconvolutional neural network which contains residual, dense deconvolution (transposed convolution), and chained residual pooing blocks.

%%%% End of Convolution-Deconvolution

\subsection{Attribute segmentation}
To the best of our knowledge, unlike lesion segmentation, there is a little literature around segmentation of attributes in dermoscopic images. Lack of annotated data is the main reason for limited number of contributions. However, after the first ISIC challenge \cite{codella2017skin}, some approaches have been proposed for this task. Kawahara et al. \cite{kawahara2018fully} proposed a segmentation model based on FCNs which incorporated a novel multi-channel Dice loss to predict 5 attributes all together. They were able to achieve first rank in ISIC2017 Task2 attributes detection.

Bissoto \textit{et al.} \cite{bissoto2018} used a UNet-like model with its encoding path  pre-trained on ImageNet data set. They also used this architecture for lesion segmentation in which they incorporated external data in their training procedures besides the original ISIC2018 challenge data set.
In the latest ISIC challenge, Sorokin \textit{et al.} \cite{sorokin2018} proposed to use Mask R-CNN model for both attributes and lesion segmentation tasks. Eric Chen \textit{et al.} \cite{chen2018attribute} also used UNet-like architecture, but they replaced the encoding path of the network with a pre-trained VGG16 model \cite{VGG}. Furthermore, two auxiliary classification branches were added to the segmentation network, one at the end of encoding path and another one at the end of decoding path to help network detecting empty attribute masks. 

Reviewed methods along with their pre-processing, post-processing, implemented augmentations and loss functions  embedded in their algorithm are summarized in Table \ref{tReview}.

\renewcommand{\tabcolsep}{3pt}

\begin{table*}[]
\centering
\tiny
\caption{Overview of different approaches applied on ISIC datasets.}

\begin{tabular}{lllllll}
\hline
\rowcolor[HTML]{FFFFFF} 
\textbf{Paper}                                & \textbf{Method}     & \textbf{Target Dataset}   & \textbf{Pre-processing}                                                                                                                                            & \textbf{Data augmentation}                                                                                                                            & \textbf{Post-processing}                                                                                            & \textbf{Loss function}                  \\ \hline
\rowcolor[HTML]{9B9B9B} 
Ahn et al. \cite{ahn2017}               & Saliency detection  & ISIC2016-Task1            & Hair and color chart inpainting                                                                                                                                    & -                                                                                                                                                     & \begin{tabular}[c]{@{}l@{}}small object removal,\\  hole filling\end{tabular}                                       & -                                       \\
\rowcolor[HTML]{FFFFFF} 
Almasani et al. \cite{almasani2018}          & Full resolution CNN & ISIC2017-Task1            & Resize to 192x256                                                                                                                                                  & \begin{tabular}[c]{@{}l@{}}Considering HSV color information\\  as separate inputs, rotation\end{tabular}                                             & -                                                                                                                   & CrossEntropy                            \\
\rowcolor[HTML]{9B9B9B} 
Berseth et al. \cite{berseth2017}           & Unet-like CNN       & ISIC2017-Task1            & Resize to 192x192,                                                                                                                                                 & \begin{tabular}[c]{@{}l@{}}Rotation, flipping, scaling,\\  elastic deformation\end{tabular}                                                           & Prediction ensembling                                                                                               & Jaccard                                 \\
\rowcolor[HTML]{FFFFFF} 
Bissoto et al. \cite{bissoto2018}           & Unet-like CNN       & ISIC2018-Task1 and Task2  & Resize to 256x256,                                                                                                                                                 & -                                                                                                                                                     & Thresholding, hole filling                                                                                          & CrossEntropy+Jaccard                    \\
\rowcolor[HTML]{9B9B9B} 
Chen et al. \cite{chen2018attribute}     & Unet-like CNN       & ISIC2018-Task2            & Resize 512x512                                                                                                                                                     & \begin{tabular}[c]{@{}l@{}}flip, rotate, scale, brightness \\ and saturation adjustment\end{tabular}                                                  & Thresholding value 0.3                                                                                              & Jaccard+CrossEntropy                    \\
Fan et al. \cite{fan2017}               & Saliency detection  & ISIC2016-Task1            & \begin{tabular}[c]{@{}l@{}}Adding HSV and\\  Lab color maps to input\end{tabular}                                                                                  & -                                                                                                                                                     & \begin{tabular}[c]{@{}l@{}}Threshoding, Hole filling, \\ spot removing\end{tabular}                                 & -                                       \\
\rowcolor[HTML]{9B9B9B} 
Galdran et al. \cite{galdran2017}           & Unet-like CNN       & ISIC2017-Task1            & shades of gray color constanc                                                                                                                                      & \begin{tabular}[c]{@{}l@{}}rotation, flipping, translation,\\  scaling, color changing\\  via illuminant estimation,\\  Gamma correction\end{tabular} & -                                                                                                                   & Jaccard                                 \\
\rowcolor[HTML]{FFFFFF} 
Jose et al. \cite{jose2019}              & Saliency detection  & ISIC2017 and ISIC26 Task2 & Resizing to fixed width of 768                                                                                                                                     & -                                                                                                                                                     & \begin{tabular}[c]{@{}l@{}}Mask refining, morphological\\  opening and closing\end{tabular}                         & -                                       \\
\rowcolor[HTML]{9B9B9B} 
Guo et al. \cite{guo2018}               & Unsupervised hybrid & ISIC2017-Task1            & -                                                                                                                                                                  & -                                                                                                                                                     & -                                                                                                                   & -                                       \\
Li et al. \cite{li2018JBHI}            & Conv-Deconv CNN     & ISIC2017 and ISIC26 Task1 & Resizeing to 400x400                                                                                                                                               & Four augmentation, not declared                                                                                                                       & -                                                                                                                   & Modified Dice                           \\
\rowcolor[HTML]{9B9B9B} 
Xu et al. \cite{xu2018}                & Unet-like CNN       & ISIC2018-Task1            & Resize to 384x512                                                                                                                                                  & Rotation, scaling, translation                                                                                                                        & Dual thresholding scheme \cite{yuan2017TMI}                                                        & CrossEntropy+Dice                       \\
\rowcolor[HTML]{FFFFFF} 
Jahanifar et al. \cite{jahanifar2018supervised}         & Saliency detection  & ISIC2017 and ISIC26 Task1 & \begin{tabular}[c]{@{}l@{}}Resize to 300x400, \\ color constancy correction,\\  hair and ruler marks inpainting\end{tabular}                                       & -                                                                                                                                                     & \begin{tabular}[c]{@{}l@{}}Thresholding,\\  objects area analysing,\end{tabular}                                    & -                                       \\
\rowcolor[HTML]{9B9B9B} 
Kawahara et al. \cite{kawahara2018fully}     & FCN-like CNN        & ISIC2017-Task1 and Task2  & Resize to 336x336                                                                                                                                                  & Rotation, flipping                                                                                                                                    & -                                                                                                                   & Dice                                    \\
\rowcolor[HTML]{FFFFFF} 
Bi et al. \cite{bi2017ISBI}            & FCN-like CNN        & ISIC2016-Task1            & \begin{tabular}[c]{@{}l@{}}Creating two distance map\\  using 2 maunally clicked position\end{tabular}                                                             & Random cropping and flipping                                                                                                                          & \begin{tabular}[c]{@{}l@{}}Thresholding, morphological dilating, \\ hole filling, small object removal\end{tabular} & CrossEntropy                            \\
\rowcolor[HTML]{9B9B9B} 
Bi et al. \cite{bi2017ISIC}            & FCN-like CNN        & ISIC2017-Task1            & Resize to have width of 500 pixels                                                                                                                                 & Random cropping and flipping                                                                                                                          & -                                                                                                                   & CrossEntropy                            \\
\rowcolor[HTML]{FFFFFF} 
Bi et al.\cite{bi2017TBME}            & FCN-like CNN        & ISIC2016-Task1            & -                                                                                                                                                                  & Random cropping and flipping                                                                                                                          & -                                                                                                                   & CrossEntropy                            \\
\rowcolor[HTML]{9B9B9B} 
Bi et al. \cite{bi2019}   & FCN-like CNN        & ISIC2017 and ISIC26 Task1 & Resize to have width of 1000 pixels                                                                                                                                & Random cropping and flipping                                                                                                                          & \begin{tabular}[c]{@{}l@{}}Thresholding, morphological dilating,\\  hole filling, small object removal\end{tabular} & CrossEntropy                            \\
\rowcolor[HTML]{FFFFFF} 
Yu et al. \cite{yu2017TMI}             & FCN-like CNN        & ISIC2016-Task1            & -                                                                                                                                                                  & \begin{tabular}[c]{@{}l@{}}Rotation, translation, \\ adding noise, cropping\end{tabular}                                                              & Thresholding                                                                                                        & Softmax CrossEntropy                    \\
\rowcolor[HTML]{9B9B9B} 
Menegola et al. \cite{menegola2017}          & Unet-like CNN       & ISIC2017-Task1            & \begin{tabular}[c]{@{}l@{}}Resizing to 128x128, and\\  ImageNet mean subtraction from input\end{tabular}                                                           & Translation, scaling, rotation                                                                                                                        & -                                                                                                                   & Dice                                    \\
\rowcolor[HTML]{FFFFFF} 
Mirkharaji et al. \cite{mirikharaji2018}       & Unet-like CNN       & ISIC2016-Task1            & \begin{tabular}[c]{@{}l@{}}Resize to 336x336,\\  normalizing by\\  mean and std\end{tabular}                                                                       & Rotation, flipping                                                                                                                                    & -                                                                                                                   & Dice                                    \\
\rowcolor[HTML]{9B9B9B} 
Navaro et al. \cite{navarro2018}           & Unsupervised hybrid & ISIC2017-Task1            & -                                                                                                                                                                  & -                                                                                                                                                     & \begin{tabular}[c]{@{}l@{}}Using Hough Transform to identify\\  and prone out hairs and bubles\end{tabular}         & -                                       \\
Aian et al.\cite{qian2018}              & FCN-like CNN        & ISIC2018-Task1            & \begin{tabular}[c]{@{}l@{}}Resizing to 512x512, Adding HSV\\  and Lab color maps to input,\\  Scaling values to (0,1) range\end{tabular}                           & \begin{tabular}[c]{@{}l@{}}Rotation, colour jittering,\\ ﬂipping, cropping, shearing\end{tabular}                                                     & \begin{tabular}[c]{@{}l@{}}Test time ensembling\\  (sample rotation)\end{tabular}                                   & Dice                                    \\
\rowcolor[HTML]{9B9B9B} 
Sarker et al. \cite{sarker2018}            & Unet-like CNN       & ISIC2017 and ISIC26 Task1 & Resize to 384x384                                                                                                                                                  & Rotation, scaling,                                                                                                                                    & -                                                                                                                   & Negative Log Likelihood+End Point Error \\
\rowcolor[HTML]{FFFFFF} 
Chen et al. \cite{chen2018}              & Unet-like CNN       & ISIC2017-Task1            & Resize 233x233                                                                                                                                                     & \begin{tabular}[c]{@{}l@{}}crop,zoom,rotate, ﬂip,\\  add gaussian noise\end{tabular}                                                                  & Threshoding, Hole filling                                                                                           & Weighted CrossEntropy                   \\
\rowcolor[HTML]{9B9B9B} 
Sorokin et al. \cite{sorokin2018}           & Mask R-CNN          & ISIC2018-Task1 and Task2  & Resize to 768x768                                                                                                                                                  & \begin{tabular}[c]{@{}l@{}}Rotation, flipping, translation,\\  gaussian blur,\\  illumination scaling\end{tabular}                                    & -                                                                                                                   & Jaccard                                 \\
Tschand et al. \cite{tschandl2019}          & Unet-like CNN       & ISIC2017-Task1            & Resize to 512x512                                                                                                                                                  & Rotation, flipping                                                                                                                                    & \begin{tabular}[c]{@{}l@{}}Thresholding with value 0.5,\\  CRF, object size filtering\end{tabular}                  & CrossEntropy-log(Jaccard)               \\
\rowcolor[HTML]{9B9B9B} 
Xiaomeng et al. \cite{xiaomeng2018}          & Unet-like CNN       & ISIC2017-Task1            & -                                                                                                                                                                  & Rotation, flipping, scaling                                                                                                                           & \begin{tabular}[c]{@{}l@{}}Thresholding with\\  value 0.5, hole filling\end{tabular}                                & CrossEntropy+MeanSquareError            \\
\rowcolor[HTML]{FFFFFF} 
He et al. \cite{he2018}                & Conv-Deconv CNN     & ISIC2017 and ISIC26 Task1 & Reducing image mean                                                                                                                                                & Rotation                                                                                                                                              & CRF                                                                                                                 & Dice                                    \\
\rowcolor[HTML]{9B9B9B} 
Xu et al. \cite{xue2018}               & Unet-like CNN       & ISIC2017-Task1            & Resize 135x180                                                                                                                                                     & \begin{tabular}[c]{@{}l@{}}flipping, color jittering,\\  cropping\end{tabular}                                                                        & Thresholding                                                                                                        & multi-scale $L_1$ distance              \\
\cite{yuan2017TMI, yuan2017} & Conv-Deconv CNN     & ISIC2017 and ISIC26 Task1 & \begin{tabular}[c]{@{}l@{}}Resize to 192x256, \\ Adding L and HSV \\c olor channels to input\end{tabular}                                                          & \begin{tabular}[c]{@{}l@{}}Rotation, flipping, translation, \\ scaling, color normalization\end{tabular}                                              & Dual thresholding scheme                                                                                            & Modified Jaccard distance               \\
\rowcolor[HTML]{9B9B9B} 
Li et al.\cite{li2018}                & FCN-like CNN        & ISIC2017-Task1 and Task2  & Center cropping, Resizing 320x320                                                                                                                                  & Rotation, flipping                                                                                                                                    & -                                                                                                                   & Weighted Softmax                        \\
\rowcolor[HTML]{FFFFFF} 
Zamani et al.\cite{tajeddin2016general}            & Unsupervised hybrid & ISIC2016-Task1            & \begin{tabular}[c]{@{}l@{}}Optimal color channel selection,\\  hair and ruler marks inpainiting,\\  dark corner detection,\\  illumination correction\end{tabular} & -                                                                                                                                                     & \begin{tabular}[c]{@{}l@{}}Morphological closing\\  and opening, hole filling, \\ border cleaning\end{tabular}      & -                                      
\end{tabular}
\label{tReview}
\end{table*}

%%%%%%%%%%%%%%%%%%%%%%%%%%%%%%%%%%%%%%%%%%%%%%%%%%%%%%%%%%%%%%%%%%%%%%%%%%%%%
\begin{figure*}[t]
	\centering
    \includegraphics[trim={0 35mm 0 35mm},clip,width=0.95\textwidth]{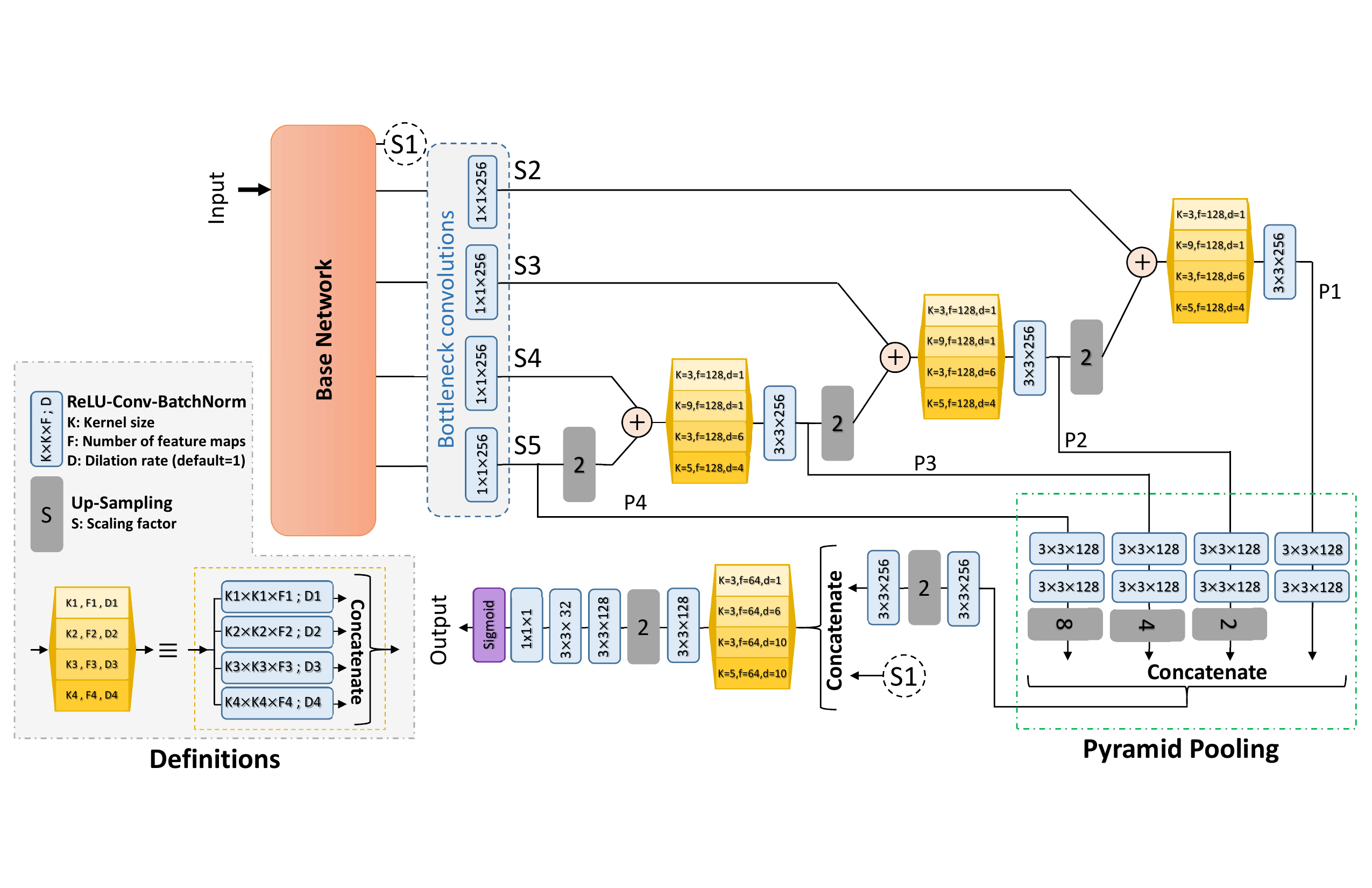}
	\caption{An overview of the proposed transfer learning based segmentation network. The proposed architecture utilizes pre-trained base networks in the encoding path, and generates the segmentation prediction map through multi-scale convolutional and  pyramid pooling blocks in the decoding path.}
	\label{fig:transfer_network}
\end{figure*}
\section{Methodolgy} \label{sec:Methodology}
In this section, we will elaborate on our approach toward Lesion and their attribute segmentation. Employment of transfer learning in our novel network architecture and proposed augmentation techniques will be covered in the following sections.
\subsection{Transfer learning}
Training deep convolutional neural networks from scratch is a challenging task due to the effects of various factors on the final outcome \cite{litjens2017survey}. These factors are designing a robust network architecture, appropriate weight initialization, setting an optimum training strategy (opting loss function, learning rate, weight decay, etc.), and a powerful hardware to support the computational complexity of CNNs. Furthermore, providing a large data set is an important requirement for CNNs to take the most advantage of their deep structures. However, in many circumstances particularly in the medical domain, annotated data is often scarce and expensive to acquire  \cite{litjens2017survey}.

To circumvent these issues, transfer learning has been widely utilized in the literature \cite{van2015transfer}. Transfer learning is about using the knowledge learned from dealing with one problem to address another (often related) problem \cite{weiss2016survey}. A common usage of transfer learning is to fine-tune powerful pre-trained deep learning models for the desired task. In this paper, we propose to use deep pre-trained models in a multi-scale framework for segmentation task.

\subsubsection{Architecture}
Our proposed segmentation networks are based on the well-known UNet architecture which comprises an encoding path to extract abstract feature maps from the input images and a decoding path to translate the extracted feature maps into output segmentation map. In order to keep the high-resolution information in the decoding path and have a more stable training process, UNet utilizes skip connections between corresponding levels of encoding and decoding path. Here, we use a variant of the UNet architecture to boost the performance on skin lesion analysis tasks.

Our proposed network benefits from a novel multi-scale convolutoinal (MSC) block which applies kernels with various size on the input. Afterwards the generated feature maps are concatenated to build the output of this block. As depicted in the "Definitions" part of Fig. \ref{fig:transfer_network}, each MSC block is built of four convolutional blocks at different scales. A convolutional block is constructed by  arranging ReLU, convolution, and batch normalization layers, sequentially. The convolution layer in the convolutional block can have different kernel sizes ($K$), number of feature maps ($F$), and dilation rates ($D$). Setting dilation rates to values more than 1 would form atrous convolutions which expands the convolution's effective filed of view while preserving resolution \cite{yu2015multi}. Our proposed MSC block can be designed arbitrarily by changing the kernel size or dilation rate of its convolutional blocks  separately, which makes it possible to form multi-scale feature extraction blocks. We set the parameters of each MSC block in our proposed network based on the location that it has been pasted, considering the level of decoding path or sizes of the MSC block input feature maps. In the following we will explain the formation of our proposed segmentation network using these building blocks.

\paragraph{Proposed multi-scale pyramid pooling network}
Firstly, for feature extraction in the encoding path of our proposed segmentation network, pre-trained models (on ImageNet data set \cite{deng2009imagenet}) are incorporated. The pre-trained networks that we utilized in this framework are 152-layered ResNet (ResNet152) \cite{ResNet}, 169-layered DensNet (DensNet169) \cite{HUANG2017DENSELY}, Xception \cite{chollet2017xception}, and Inception-ResNet v2 (ResNetV2) \cite{szegedy2017inception}. These models are referred to as the base model in our proposed segmentation framework and each of them has their own technical details which explaining them is beyond the extend of the this paper. The important note about all of these base networks is that they all reduce the size of the input image or its feature maps in order of five times i.e., either max pooling or strided convolution layers were utilized five times throughout all base networks’ architecture. Therefore, feature maps through the decoding path of our segmentation framework must be up-sampled exactly five times in order to construct an output prediction map with the same size as the input. 

The schematic architecture of the proposed transfer learning based segmentation network is depicted in Fig. \ref{fig:transfer_network}. In this figure, “Base Network” block can be replaced by any model from previously mentioned networks (resnet152, densenet169, xception, or resnetv2).
Essential feature maps are extracted from base model  to construct our proposed  segmentation network (labeled as S1, S2, S3, S4, and S5 in Fig. \ref{fig:transfer_network}). If we consider the base network in different levels (scales) where transitions between scales are made by max pooling or strided convolution, S1 to S5 tags can be taken regarding to the following instructions: Generally, S1 would be the output from the first max pooling or strided convolution layer (from the first level) of the base network. Clearly, size of feature maps that are tagged with S1 label would be half of the input image. On the other hand, tags S2 to S4 are pointing to the feature maps that are extracted from the last non-strided convolutional layers of subsequent levels, right before  max pooling layer. The last extracted feature map, labeled as S5, would be the output of the last convolutional layer i.e., the output of the base network without including top layers (pooling and dense layers).

Having feature maps extracted from five different levels of the base model, we can build the decoding path of our segmentation network in a pyramid pooling manner.
Before going any further, all of the S1 to S5 feature maps are passed through $1\times1\times256$ bottleneck convolutional blocks  to make the number of extracted feature maps in all levels equal to 256. This operation would enable us to merge (add) resized feature maps extracted from different levels and construct skip connections. Moreover, this would reduce the number of overall network parameters and computational complexity, allowing  to use larger batch sizes in the training procedure. 
Outputs of the last convolutional layer from the base network are extracted and up-sampled followed by MSC and $3\times3\times256$ convolutional blocks in each level of the decoding path. As you can see in Fig. \ref{fig:transfer_network}, parameters of MSC blocks are different in each level of decoding path due to different size of feature maps in those levels. In the higher levels, we set the MSC parameters to make its convolutional blocks capture bigger field of views by either increasing the kernel size or increasing the dilation rate.  Note that unlike the original UNet, feature maps between encoding and decoding paths are merged by "add" operators.

Feature maps from different levels of decoding path are then gathered in a pyramid feature pooling manner: from each level, the feature maps before up-sampling layers (P2 $ \sim $ P4) and the output from the last convolutional layer (P1) are extracted. Then, all of them are passed through two $3\times3\times128$ convolutional blocks. Each of P2, P3, and P4 lanes are then up-samples accordingly to make the size of their feature maps equal to the size of lane P1 feature maps. Afterward, feature maps from all lanes are concatenated to form the output of the pyramid feature pooling block which is up-sampled and adjusted through convolutional blocks and $2\times$ up-sampling layer. Then, it is concatenated with the biggest feature maps extracted from the base network (S1). The concatenation output is abstracted through MSC and $3\times3\times128$ convolutional blocks and up-sampled once again to generate feature maps with the same size as the network input. Two more $3\times3$ convolutional blocks with 128 and 32 feature map, respectively, are utilized to reduce the number of feature maps as we reach the end of the network. The final output (segmentation map) is constructed by applying a  $1\times1\times1$ convolution followed by a sigmoid activation layer.

\paragraph{DeepLab v3}
Apart from UNet-like architectures, we also incorporated DeepLabV3 \cite{chen2018encoder} network for lesion segmentation task. DeeplabV3 is an atrous convolution enhanced network architecture, originally designed for semantic segmentation. The DeepLabV3 uses Xception \cite{chollet2017xception} architecture as its base network and at the end of its encoding path, an atrous spatial pyramid pooling scheme has been incorporated \cite{chen2018encoder}, which is applying dilated convolutions with different rates. DeepLabV3 does not use skip connections nor multi-level up-sampling in its decoding path, Instead the feature maps are resized through bilinear interpolation followed by convolutions. The only modification that we have made to the DeepLabV3 network to apply for our problem, is changing the last convolutional layer to get only one segmentation map in the output. Similar to previously mentioned  networks, the output layer consists of a 1x1 convolution with sigmoid activation. It is worth mentioning that our pre-trained weights for DeepLabV3, was originally obtained based on Pascal data set \cite{Everingham10}.

\subsubsection{Loss Function}
A modified Jaccard loss function \cite{yuan2017TMI} is used for lesion and attribute segmentation:
\begin{equation}
Loss = (1 - \frac{{\sum {{y_t}{y_p} + \alpha } }}{{\sum {y_t^2 + \sum {y_p^2 - \sum {{y_t}{y_p} + \beta } } } }})
\end{equation}
In which, $y_p$ is network prediction and $y_t$ is ground truth segmentation. This modified Jaccard loss function makes the training process converge faster and more stably. It is also robust against the pixels’ imbalanced class population and able to predict a smooth segmentation output \cite{yuan2017TMI}. Smoothing parameters $\alpha$ and $\beta$ are set differently and specifically for each lesion or attribute segmentation task.

\subsubsection{Preprocessing}
Images are preprocessed according to the type of the base model that has been used. For each base model, different preprocessing schemes were applied when they were trained on the ImageNet data set. Therefore, we incorporate the same preprocessing procedures originally deployed for each individual base model.\\
If the base network is one of the ResNet152 or  ResNetV2 the preprocessing procedure comprises subtracting the channel-wise mean of the ImageNet data set from each image. If the base model is DensNet169, ImageNet mean image is subtracted from each input and the result is divided by  dataset standard deviation . For the Xception based model, image values are re-scaled between (-1,1) and for the DeeplLabV3, input images are re-scaled between  (0, 1). 
Additionally,  for both lesion and attribute segmentation tasks, images are resized to 384$\times$576.

\subsection{Data augmentation}

To eliminate the effect of the small data set, an extensive set of augmentations has been incorporated. All augmentations were carefully designed to simulate all the possible variations and transformation present in dermoscopic images.
Explaining the details of all incorporated augmentation routines is beyond the scope of this manuscript. However, a curated list of employed augmentations are: Random vertical and horizontal flips, random rotation in the range of 0 to 40 degrees, image zooming with a random  scale between 0.7 to 1.3, random image translation, random image shearing to the extent of 0.3, random shift in color channels (color jittering) up to 40 units, scaling image intensities with a random scale in the range of 0.7 to 1.3, and adding noises from different random distributions and types (Gaussian, Speckle, Salt \& Pepper).

Furthermore, we have introduced four new routines to augment dermoscopic images, which can imitate dermoscopic images variations more realistically. These new augmentations are described below. Note that all of the following and  above-mentioned augmentations are applied "on the fly" during training.

\subsubsection{Contrast adjustment}
Training the network with contrast adjusted images can help dealing better with cases that have been captured with poor quality. In order to simulate realistic contrast situations, we can stretch and shrink the image contrast using the linear intensity scaling (LIS) technique as follow:
\begin{equation}
{{\bf{X}}_{{\rm{out}}}} = ({{\bf{X}}_{{\rm{in}}}} - {I_{{\rm{low}}}})\left( {\frac{{{O_{{\rm{high}}}} - {O_{{\rm{low}}}}}}{{{I_{{\rm{high}}}} - {I_{{\rm{low}}}}}}} \right) + {O_{{\rm{low}}}}.
\label{eq_contrast}
\end{equation}

In the above equation,  ${\bf{X}}_{{\rm{out}}}$ and $\bf{X}_{\rm{in}}$ are output and input images, respectively, and $ (I_{{\rm{low}}}, I_{{\rm{high}}})$ are the lower and upper limits for pixel values from the input image which we desire to map  onto $ (O_{{\rm{low}}}, O_{{\rm{high}}})$ in the output image.
\paragraph{Shrinking/lowering the contrast}
In order to decrease image contrast, we set the $ (I_{{\rm{low}}}, I_{{\rm{high}}})$ limit values equal to input image minimum and maximum values i.e., $ (\min ({{\bf{X}}_{{\rm{in}}}}),\max ({{\bf{X}}_{{\rm{in}}}}))$. Corresponding lower and upper bounds in the output images are then obtained randomly based on the image maximum and minimum intensity values:
\begin{equation}
\begin{array}{l}
{O_{{\rm{low}}}} = \min ({{\bf{X}}_{{\rm{in}}}}) + \alpha max({{\bf{X}}_{{\rm{in}}}}),\\
{O_{{\rm{high}}}} = \max ({{\bf{X}}_{{\rm{in}}}}) - \beta max({{\bf{X}}_{{\rm{in}}}}),
\end{array}
\end{equation}
In which, $\alpha$ and $\beta$ are scalars in the range of $(0.1,0.3)$,  randomly drawn from a uniform distribution.

\paragraph{Stretching/increasing the contrast}
For increasing image contrast, $ I_{{\rm{low}}}$ and $ I_{{\rm{high}}}$ are set to be 2\textsuperscript{nd} and 98\textsuperscript{th} percentiles of the input image values, respectively. On the other hand, output intensity bounds, $ (O_{{\rm{low}}}, O_{{\rm{high}}})$, are set to the minimum and maximum possible values in a standard 8-bit image i.e., $(0,255)$.

In the left column of Fig. \ref{fig:aug_csi}, a sample dermoscopic image alongside its contrast adjusted  versions are depicted. In our implementation, we applied the contrast stretching or shrinking with equal probability during the training.

 \subsubsection{Sharpness adjustment}
 In dermoscopy, having out-of-focus images is a prevailing artefact. To imitate this situation, we proposed to blur the input image by applying a Gaussian filter which is quite popular augmentation technique. To this end, a two-dimensional Gaussian kernel with standard deviation of $\sigma$ and window size of $w = 2\left\lfloor {4\sigma  + 0.5} \right\rfloor  + 1$ has been used. In the current implementation, $\sigma$ is a randomly drawn number in the range of $(0.6,1.2)$.
However, in this research we proposed to apply the reverse procedure on dermoscopic images as a new augmentation technique , therefore we can also have sharpen version of captured images in the training . This augmentation will help network detecting subtle details in the image, like dermoscopic attributes. To increase image sharpness, we incorporated unsharp masking technique. Unsharp mask (UM) is constructed by scaling the negative of blurred input image. The unsharp mask is then combined with input image to enhance its sharpness. Our implementation of unsharp masking is as below:
\begin{equation}
{{\bf{X}}_{{\rm{sharp}}}} = (1 + a){{\bf{X}}_{{\rm{in}}}} - a({\bf{X}} * {\mathop{\rm g}\nolimits} (\sigma )),
\label{eq_unsharp}
\end{equation}
In which, ${\mathop{\rm g}\nolimits} (\sigma )$ is a Gaussian kernel with standard deviation of $\sigma$ and {$*$}  indicate convolution operation. Besides $\sigma$, parameter $a$ is for controlling the amount of sharpening. We set the parameters to constant values of ${\sigma}=2$ and $a=1$. The middle column of Fig. \ref{fig:aug_csi} illustrates smoothed and sharpened versions of a dermoscopic images as an augmented inputs for the network.
 
 \subsubsection{Disturbing illumination uniformity}
\begin{figure}[t]
%\centering
\includegraphics[width=1.0\columnwidth]{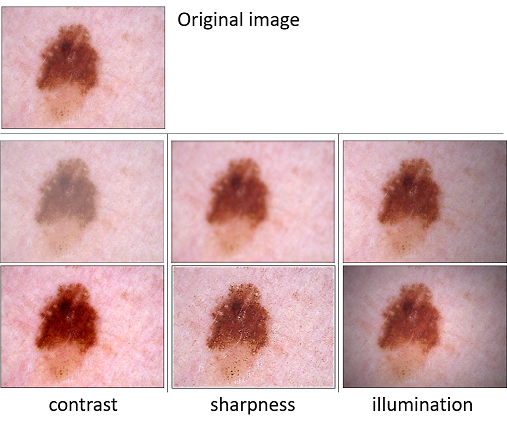}
	\caption{Three augmentation techniques applied on dermoscopic images.From left to right, first column shows image contrast variation using linear intensity scaling. Second column demonstrated changing image sharpness using Gaussian filtering and unsharp masking and in the last column image illumination uniformity is disturbed by axial and radial illumination gradient maps.}
\label{fig:aug_csi}
\end{figure}

Due to inappropriate imaging condition or poor lighting, dermoscopic images may show uneven illumination in their background. This effect may interfere with lesion or attributes segmentation in severe cases. So, we propose to synthetically perturb background illumination . For this end, we multiplied the input image by a randomly generated illumination gradient map (IGM). The gradient map can be either axial (horizontal or vertical) or radial.
\paragraph{Axial IGM}
Assume we have an input image with $R$ rows and $C$ columns.  To generate an horizontal (vertical) IGM, two scalars are randomly selected for $high$ and $low$ illumination scale of the gradient map. We sample the $low$ scale from a uniform  distribution in range of $(0.6,0.75)$ and, similarly, the $high$ scale from range of $(1.1,1.3)$. Then, $C$ evenly spaced numbers are generated from the $\left[ {low,high} \right]$ interval. Finally, to construct the IGM, that gradient line is repeated $\# R$ ($\# C$) times vertically (horizontally).
\paragraph{Radial IGM}
To generate radial IGMs, first, we must select a random position at the center of radial IGM, like $ {x_{\rm{c}}} \in (0,C)$  and $ {y_{\rm{c}}} \in (0,R)$, in which $R$ and $C$ are number of rows and columns in the input image, respectively. Then, Euclidean distance (${\bf{D}}(x,y)$) of every other positions (pixels) in the image to this position is calculated (${\bf{D}}(x,y) = \sqrt {{{(x - {x_{\rm{c}}})}^2} + {{(y - {y_{\rm{c}}})}^2}}$). To achieve plausible multiplier factors in the radial IGM, the obtained distance function is inverted, ${\bf{\tilde D}} = \max ({\bf{D}}) - {\bf{D}}$, and then adjusted using the linear intensity scaling technique (eq. \ref{eq_contrast}). In the LIS formula, $(I_{{\rm{low}}}, I_{{\rm{high}}})$ are set to $(min({\bf{\tilde D}}),max({\bf{\tilde D}}))$ and $(O_{{\rm{low}}}, O_{{\rm{high}}})$ are set to $(low,high)$ factors (sampled randomly similar to previous paragraph). Implementing of this augmentation technique on simple image is illustrated in the right-most column of Fig. \ref{fig:aug_csi} , in which illumination of the upper and the lower images have been disturbed using axial and radial IGMs, respectively.

 \subsubsection{Hair occlusion}
 One of the most common deficiencies in dermoscopic images is hair occlusion. with higher presence of hairs in the image, lesion segmentation, attribute detection, and disease classification tasks would be harder to correctly achieve using CAD systems. This phenomenon may confuse dermatologist in decision making too. In current available dermoscopic data sets, some images are occluded with hair. However, in order to make the network more robust against this deficiency, we propose to increase the occurrence of  hair occluded images in the training phase by synthetically adding hairs to the input image.
To achive this task in real-time, a randomly selected hair map is multiplied by the input image. We synthetized hair maps using HairSim toolbox\footnote{available: www.cs.sfu.ca/∼hamarneh/software/hairsim}. HairSim is a hair simulation toolbox in Matlab developed by Mirzaalian et al. \cite{mirzaalian2014hair}. HairSim create random curves as the medial axis of each synthetized hair, then it thickens the curves by dilation operation which incorporates a size-varying disk shape structuring element, and finally, coloring each hair line with specific shades \cite{mirzaalian2014hair}.

\begin{figure}[t]
%\centering
\includegraphics[width=1.0\columnwidth]{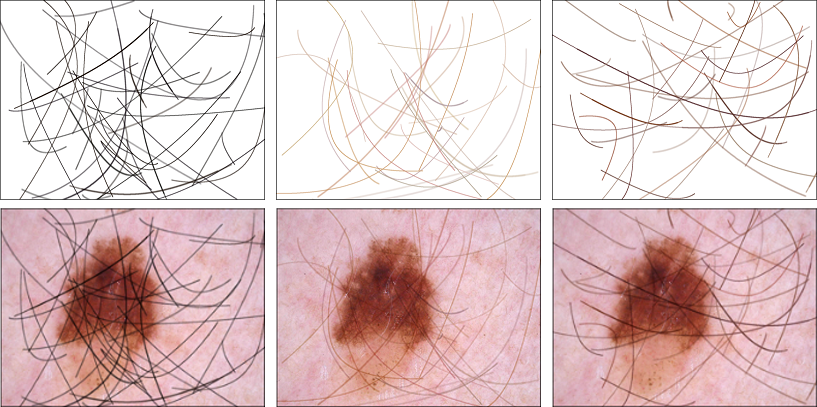}
	\caption{Synthetically polluting dermoscopic images with hairs. First row shows three random generated hair maps with different colors and geometric characteristics and the second row illustrates dermoscopic images occluded with generated hair maps.}
\label{fig:aug_hair}
\end{figure}

 In the current work, we modified and harness the HairSim toolbox to automatically pollute a series of white backgrounds with synthetized hair lines, which have specific random features. These hair polluted images serves as hair maps in our augmentation framework. To make hair maps as realistic as possible, we reviewed the available data sets and tuned our hair map generation procedure. Number of synthetized hairs, their spread, thickness, and curliness parameters were selected from non-uniform and plausible distributions, which were obtained heuristically. Also, twenty distinct color profiles were used to colorize synthetized hairs, which comprised black, dark-gray, dark-brown, light-brown, light-gray, dark-gold, light-gold and white colors and their combination. With having hair maps generated, we can use them to augment dermoscopic images during the training i.e., polluting the input image with hairs by multiplying a randomly selected hair map into the input image. Fig. \ref{fig:aug_hair} shows some generated hair maps and their resultant hair polluted images.

\subsection{Lesion Segmentation}
\subsubsection{Training procedure} \label{subsec:training}
All decoding layers of the proposed segmentation network are initialized using “uniform Glorot” \cite{ glorot2010understanding } initialization method. Loss optimization has been carried out using AMSGrad variant of Adam optimizer \cite{ reddi2018convergence } with the learning rate and weighs decay parameters equal to ${10^{ - 4}}$ for all layers. A 5-fold cross-validation scheme with 70 epochs for each fold and batch size of 16 was utilized.  The best performing model on the validation subset is saved during the training process based on the loss value. For lesion segmentation, we set the  coefficient $\alpha=1$ and $\beta = 1$  .

\subsubsection{Post processing} \label{postprocess}
For acquiring the final binary mask, first, the lesion marker is constructed by applying threshold (${T_H}$) on the prediction map. Lesion mask  is also obtained through thresholding the prediction map with upper value than (${T_L}$). Afterward, lesion marker is refined to consist only an object with the biggest area. Consequently, the segmentation mask of the lesion is constructed through binary morphological reconstruction. %\cite{gonzalez2010morphological}:
%\begin{equation}
%\begin{array}{l}
%Output = imreconstruct(Lesion_{marker},%Lesion_{mask})
%\end{array}
%\end{equation}
The threshold values for ${T_H}$ and ${T_L}$ are chosen based on a grid search on different values in the ranges ${T_H} \in [0.8,0.9,0.95,0.975,0.99,0.995,0.996]$ and ${T_L} \in [0.3:0.05:0.85]$.

\subsection{Lesion attributes segmentation}
A few attempts in the literature have been made to address the problem of lesions' attribute segmentation. Unlike \cite{kawahara2018fully} which tried to train a network predicting all 5 attribute together, in this paper, we proposed to use separate models to segment each attribute. 
%In other words, for detecting each attribute, we propose to train a specific model.
Models for these tasks are the same as the models we introduced in the previous section for segmenting the lesion mask.

\subsubsection{Data Subsampling}
An important challenge in attribute segmentation task is that the population of different attributes in the available data sets are very uneven. For instance, pigment network is an attribute that is present in the most of dermoscopic images whereas streaks are very rare attributes. As we mentioned in the previous section, we address all attributes segmentation as a separate binary segmentation task. To deal with the unbalanced class in that binary segmentation task (which is very crucial when we are looking for attributes like Streaks that has a considerably smaller share of the data set), we propose to use subsampling of negative class.

For clarification, an example of subsampling procedure for streak attribute is given in the following. Out of 2594 images in the ISIC2018 data set \cite{Tschandl2018_HAM10000}, 100 images contain Streaks attribute. To deal with imbalanced class population, we randomly sample 100 images from negative class (remaining 2494 image without Streaks). Although the number of data is equally divided between the two classes, many training data are ignored for this purpose. Nevertheless, the class population adjustment leads to having a sensible prediction map.

\subsubsection{Training procedure}
The training procedure for segmenting each attribute is done just like the process we explained for lesion segmentation (section \ref{subsec:training}), However, to prevent network to be biased toward negative class and to clearly predict all pixels to 0 value, we set smoothing coefficients $\alpha=0$ and  $\beta=1$ for attribute detection as recommended in \cite{kawahara2018fully}.

\subsubsection{Post-processing}
Post-processing procedures for attributes consists of restricting the prediction map to the lesion area (by multiplying the attribute predicted map into lesion segmentation mask), finding the high and low thresholding values (${T_H},{T_L}$) to construct plausible $Attribute_{marker}$ and $Attribute_{mask}$, and generating the final attribute segmentation using morphological reconstruction. Note that unlike lesion segmentation task, here we do not remove any objects from the constructed marker by area analysis.

\subsection{Test time augmentation and prediction ensembling}
It has been shown that ensembling techniques can elevate the performance considerably \cite{codella2017deep}, therefore we tried to use ensembling of predictions as much as possible. Various test time augmentations are applied and then their outputs are merged (averaging). These ensembled predictions are then going under post-processing routines.

For all segmentation models, test images are predicted using all trained weights in 5 folds experiments. In each fold, the original image is augmented with four different transformations:  horizontal flip, vertical flip, contrast enhancing  of a $90^\circ$  rotated version of the image, and  sharpening the image. Hence, to construct the final prediction from a single model in our proposed approach, 25 predictions are acquired and averaged. % Average of these 25 predictions is considered as the output of that model for a single image.%
Moreover, the final prediction maps from different models are merged together to obtain the ultimate ensemble prediction.

\section{Experimental setups and results} \label{sec:setups}

\begin{table*}[h]
\centering
\caption{Evaluation metrics obtained from cross-validation experiments on ISIC2018 Task1
training set using different base models in the proposed lesion segmentation framework}
\begin{tabular}{lccccccc}
\hline \hline
\textbf{Base Network} & (${T_H,T_L}$)        & \textbf{TJI}& \textbf{JSI} & \textbf{DSC}   &\textbf{ ACC }&\textbf{ SEN} & \textbf{SPC} \\ \hline
DenseNet169  & (0.8, 0.5)  & 79.96              & 83.84  & 90.40 & 96.58   & 91.35      & 97.12       \\ 
ResNet151    & (0.8, 35)   & 79.91              & 83.68  & 90.32 & 96.48   & 92.00      & 96.65      \\ 
ResNetv2     & (0.8, 0.4)  & 80.34              & 84.02  & 90.49 & 96.58   & 91.80      & 97.17      \\ 
Xception     & (0.8, 0.5)  & 79.93              & 83.86  & 90.38 & 96.56   & 91.23      & 97.30      \\ 
DeepLab v3   & (0.8, 0.5)  & 79.32              & 83.38  & 90.30 & 96.55   & 91.15      & 97.32      \\ 
Ensemble     & (0.8, 0.45) & 81.42              & 84.68  & 91.00 & 96.80   & 92.09      & 97.32      \\ \hline \hline
\end{tabular}
\label{tSegTrain}
\end{table*}

\begin{table*}[]
\centering
\caption{Evaluation metrics resulted from ISIC2018 Task1, lesion segmentation, testing set}
\begin{threeparttable}
  \centering
\begin{tabular}{llll}
\hline \hline
Rank & User           & Method Description\tnote{$\dagger$}                                       & TJI \\ \hline
1    & Chengyao Qian  & MAskRcnn2+segmentation {[}Qian2018\_ISIC{]}              & 80.2       \\
2    & Young Seok     & ensemble with CRF\_v3\tnote{*1}                                & 79.9       \\
3    & ji YuanFeng    & Automatic Skin Lesion Segmentation by DCNN\tnote{*2}         & 79.9       \\
4    & Yuan Xue       & Skin Lesion Segmentation with Adversarial Learning\tnote{*3}   & 79.8       \\
5    & Navid Alemi    & Proposed Method-ensemble prediction-(Th,TI0)=(0.80,0.65) & 79.6       \\
6    & Chengyao Qian  & MaskRcnn2+segmentation2{[}Qian2018\_ISIC{]}              & 79.4       \\
7    & Young Seok     & ensemble method with CRF\tnote{*1}                            & 79.4       \\
8    & Young Seok     & ensemble method with CRFV2\tnote{*1}                           & 79.4       \\
9    & Miguel Molina  & SR FCN Init2\tnote{*4}                                        & 78.8       \\
10   & Navid Alemi    & Proposed Method-Ensemble Prediction-(Th,TI)=(0.8,0.5)    & 78.4       \\
11   & Miguel Molina  & SR FCN ensemble Inits\tnote{*4}                                & 78.1       \\
12   & Andrey Sorokin & Mask-RCNN with SGD optimizer {[}sorokin2018{]}           & 77.9       \\ \hline \hline
\end{tabular}
\begin{tablenotes}
    \item [$\dagger$] Results in this table are directly adopted from the ISIC2018 challenge Task 1 test leaderboard. %(https://challenge.kitware.com/\#phase/5b1c193356357d41064da2ec). 
    Entries marked with "*" symbol were not accompanied by a cite-able article. However, their submission report titles are listed below:
        \item[*1] Team HolidayBurned at ISIC CHALLENGE 2018
        \item[*2] Automatic Skin Lesion Segmentation by Feature Aggregation Convolutional Neural Network
        \item[*3] Automatic Skin Lesion Analysis with Deep Networks
        \item[*4] An Elliptical Shape-Regularized Convolutional Neural Network for Skin Lesion Segmentation
    \end{tablenotes}
\end{threeparttable}%
\label{tISIC2018_segTest}
\end{table*}

\subsection{Data sets}
Experimental data sets for the current research have been adopted from “The International Skin Imaging Collaboration” (ISIC) archive. ISIC has started “Melanoma Project” in order to help early recognition of melanoma and reducing its mortality rate. For the same purpose, so far ISIC has organized three grand challenges on skin lesion segmentation, detection of clinical diagnostic patterns, and disease classification under the title of “ISIC: Skin Lesion Analysis Towards Melanoma Detection“ \cite{gutman2016skin,codella2017skin,Tschandl2018_HAM10000}. Based on the year that challenges have been held, they are named as ISIC2016, ISIC2017, and ISIC2018. Relatively, we would use the same abbreviations for the data set naming in the rest of the paper. Each data set comprises a different number of training and testing images for different tasks. Since we are working on skin lesion segmentation and clinical attributes detection tasks, in the following paragraphs we would only describe their related statistics from the ISIC2016, ISIC2017, and ISIC2018 data sets.

ISIC2018 data set’s training part comprises 2594 alongside with ground truth segmentation of lesion boundary (Task 1) and five dermoscopic attributes (Task 2: detection of pigment network, negative network, globules, milia-like cysts, and streaks). Apart from the training set, ISIC2018 has a test set of 1000 dermoscopic images \cite{codella2017skin,Tschandl2018_HAM10000}. By the date that this paper is written, the ground truth segmentation for test set is not released for public and methods evaluation is done through the challenge website, therefore, the reported results for ISIC2018 test sets in this paper, either lesion segmentation or attribute detection tasks, are directly adopted from there.

ISIC2017 Task 1 training set comprises 2000 annotated dermoscopic images in JPEG format accompanied by ground truth segmentations in PNG format. The test set for ISIC2017 has 600 images with their ground truth segmentation available to the public \cite{codella2017skin}. For Task2 of the challenge, ISIC2017 have the ground truth of dermoscopic attributes available for training and testing sett.Since they organized in a super-pixel labeling format rather than image masks .we didn’t carry out any attribute detection on ISIC2017 Part 2: Lesion Dermoscopic Feature Extraction data set.

 ISIC2016 consists of 900 training and 379 testing images with all images having ground truth segmentations available for the public. In comparison to ISIC2017 and ISIC2018 is a smaller data set, but there are a considerable literature and research reports on it, hence, we decided to consider it in our experiments. Like ISIC2017, ISIC2016 introduced separate training and test sets for "globules" and "streaks" dermoscopic features detection task, but we decided not to include it in our experiments due to lack of literature/research background.

\subsection{Evaluation metrics}
Evaluation metrics for lesion segmentation tasks are threshold Jaccard index (TJI), Jaccard similarity index (JSI), sensitivity (SEN), specificity (SPC), accuracy (ACC), and Dice similarity coefficient (DSC) \cite{codella2017skin}:
\begin{itemize}
    \item $ JSI = {{\left| {{\bf{G}} \cap {\bf{P}}} \right|} \mathord{\left/
 {\vphantom {{\left| {{\bf{G}} \cap {\bf{P}}} \right|} {\left| {{\bf{G}} \cup {\bf{P}}} \right|}}} \right.
 \kern-\nulldelimiterspace} {\left| {{\bf{G}} \cup {\bf{P}}} \right|}} $
 
    \item $ DSC = 2{{\left| {{\bf{G}} \cap {\bf{P}}} \right|} \mathord{\left/
 {\vphantom {{\left| {{\bf{G}} \cap {\bf{P}}} \right|} {\left( {\left| {\bf{G}} \right| + \left| {\bf{P}} \right|} \right)}}} \right.
 \kern-\nulldelimiterspace} {\left( {\left| {\bf{G}} \right| + \left| {\bf{P}} \right|} \right)}} $
 
   \item $ ACC = {{\left( {{\rm{TP + TN}}} \right)} \mathord{\left/
 {\vphantom {{\left( {{\rm{TP + TN}}} \right)} {\left( {{\rm{TP + FP + TN + FN}}} \right)}}} \right.
 \kern-\nulldelimiterspace} {\left( {{\rm{TP + FP + TN + FN}}} \right)}} $
 
    \item $ SEN = {{{\rm{TP}}} \mathord{\left/
 {\vphantom {{{\rm{TP}}} {\left( {{\rm{TP + FN}}} \right)}}} \right.
 \kern-\nulldelimiterspace} {\left( {{\rm{TP + FN}}} \right)}} $
 
    \item $ SPC = {{{\rm{TN}}} \mathord{\left/
 {\vphantom {{{\rm{TN}}} {\left( {{\rm{TN + FP}}} \right)}}} \right.
 \kern-\nulldelimiterspace} {\left( {{\rm{TN + FP}}} \right)}} $
\end{itemize}
 %In the above equations, $\left| . \right|$ operator denotes the number of non-zero elements in each of ground truth  (${\bf{G}}$), or predicted segmentation (${\bf{P}}$).%
 ACC, SEN, and SPC  are metrics that indicate the goodness of segmentation by evaluating the classification of each pixel into positive or negative clasees. TP (TN) is equal to the number of pixels in the predicted segmentation that are correctly labeled as positive (negative), and contrariwise, FP (FN) is the number of pixels that were incorrectly labeled as positive (negative). TJI is the thresholded version of Jaccard similarity index in which if the value of JSI is below 0.65, the TJI would be, in other words \cite{tschandl2019}:

\begin{itemize}
    \item $ TJI = \left\{ {\begin{array}{*{20}{c}}
{JSI}&{{\rm{if }}JSI \ge {\rm{0}}{\rm{.65}}}\\
0&{{\rm{if }}JSI \le {\rm{0}}{\rm{.65}}}
\end{array}} \right. $
\end{itemize}

In ISIC2016 and ISIC2017, the selected metric to rank participant for lesion segmentation task was Jaccard index but for ISIC2018 it was threshold Jaccard index (averaged over all test images).

For the second task (attribute segmentation), each attribute is evaluated separately in a way that all output segmentation maps from all test image are first concatenated together to form a multi-channel output with a specific size ($256\times256\times${number of test images}), similar procedure is done for their relative ground truth masks, and then the similarity between the multi-channeled ground truth and prediction is evaluated using the Jaccard index and Dice coefficient. We name these metrics in the result tables as cumulative Jaccard (C-JSI) and cumulative Dice (C-DSC). The final evaluation of a method is drawn out by concatenating all 5 different attributes predictions from all test images.

Note that all of the above-mentioned metric values are in the range of (0,1), but for clarity, we reported them as performance percentages in the range of (0,100).

%\begin{table}[]
%\centering
%\caption{ Evaluation metrics obtained from ISIC2018-Task1 validation set using different base models in the proposed segmentation framework.}
%\resizebox{.48\textwidth}{15mm}{%
%\begin{tabular}{lcc}
%\hline \hline
%\textbf{Base Network} & \textbf{${T_H,T_L}$} & \textbf{Thresholded Jaccard} \\ \hline
%DenseNet169  & (0.8, 0.65) & 0.826               \\ 
%ResNet151    & (0.8, 0.65) & 0.831               \\ 
%ResNetv2     & (0.8, 0.65) & 0.791               \\ 
%Xception     & (0.8, 05)   & 0.799               \\ 
%DeepLab v3   & (0.8, 0.65)  & 0.825               \\ 
%Ensemble     & (0.8, 0.65) & 0.820               \\ \hline \hline
%\end{tabular}
%}
%\label{tSegVal}
%\end{table}

\subsection{Implementation details}
Model training, validation and prediction have been implemented using Keras library  \cite{chollet2015keras} with Tensorflow backend. Grid search on thresholds, prediction ensembling, post-processing have been done with the help of Mathwork Matlab 2016b. All experiments were done on an Intel Core i9 machine equipped with 128 GB of RAM and four GPUs (Nvidia Geforce GTX 1080 Ti) running on windows. Moreover, we have adopted the DeepLabV3 implementation from \url{https://github.com/bonlime/keras-deeplab-v3-plus}.

%%%%%%%%%%%%%%----------------------------------------

\subsection{Lesion segmentation results}
In order to assess our lesion segmentation framework in action and show how post-processing thresholds have been selected, we reported the results on the training set only for ISIC2018 data set, which has the biggest training population among the others. For ISIC2016 and ISIC2017 data sets, however, we suffice to report the results only on the test sets. The following subsections describe the results on these data sets quantitatively and qualitatively.

\subsubsection{ISIC2018}
Table \ref{tSegTrain} shows the lesion segmentation evaluation results extracted from 5 fold cross-validation experiments on 2549 training images of ISIC2018 data set using different base models. These performance metrics are derived from the final segmentation masks, which are constructed by post-processing each network prediction. Main steps the post-processing task, as described in \ref{postprocess}, is two thresholding operation.
%The optimal values for high threshold, $T_H$, and low threshold, $T_L$ are selected by performing a threshold grid search on the outputs from the cross-validation experiment.%
The optimal threshold values for each base network and the ensemble predictions are reported in the second column of Table \ref{tSegTrain} as a (${T_H,T_L}$) tuple. These selected thresholds are then used to construct the final segmentation for the test sets. Please keep in mind that the procedure of selecting the optimal thresholds from training set is repeated for all data sets and tasks, but for the sake of brevity we do not express them for other experiments.

It is inferred from Table \ref{tSegTrain} that "resnetv2" outperforms other base models with TJI value of 80.34. To elevate segmentation results, we ensemble predictions from different models before post-processing. Evaluation metrics derived from ensemble predictions are better than every single model performance, showing about 1 percent advantage over the best performing single model on the training set (resnetv2) based on TJI metric.

As mentioned before, the ground truth segmentations for the ISIC2018 test set are not released by the organizer. Therefore, we restricted to only report the TJI metric, as the ISIC2018 challenge leader board\footnotemark provides only this metric. Evaluation metric resulted from the applying the proposed segmentation method on the ISIC2018 test set is reported in Table \ref{tISIC2018_segTest}. Our proposed method is submitted under the “Navid Alemi” user name in the leader board. Our submission was able to achieve the 5th rank in the leader board among 112 different submitted competitors. As you can see in Table \ref{tISIC2018_segTest}, our method is very competitive to other high rankers as the result are marginally close the first three ranks with only 0.2\% difference in performance (based on the absolute value of TJI metric, only 0.003 lower than tied second and third positions). However, this results show the power of our proposed segmentation algorithm, besides, a small difference of obtained metric for training and test data (1.8\% in performance or 0.018 difference of TJI values in train and test) represents the generalization of the proposed segmentation method.

\begin{figure*}[t]
	\centering
    \includegraphics[width=0.90\textwidth]{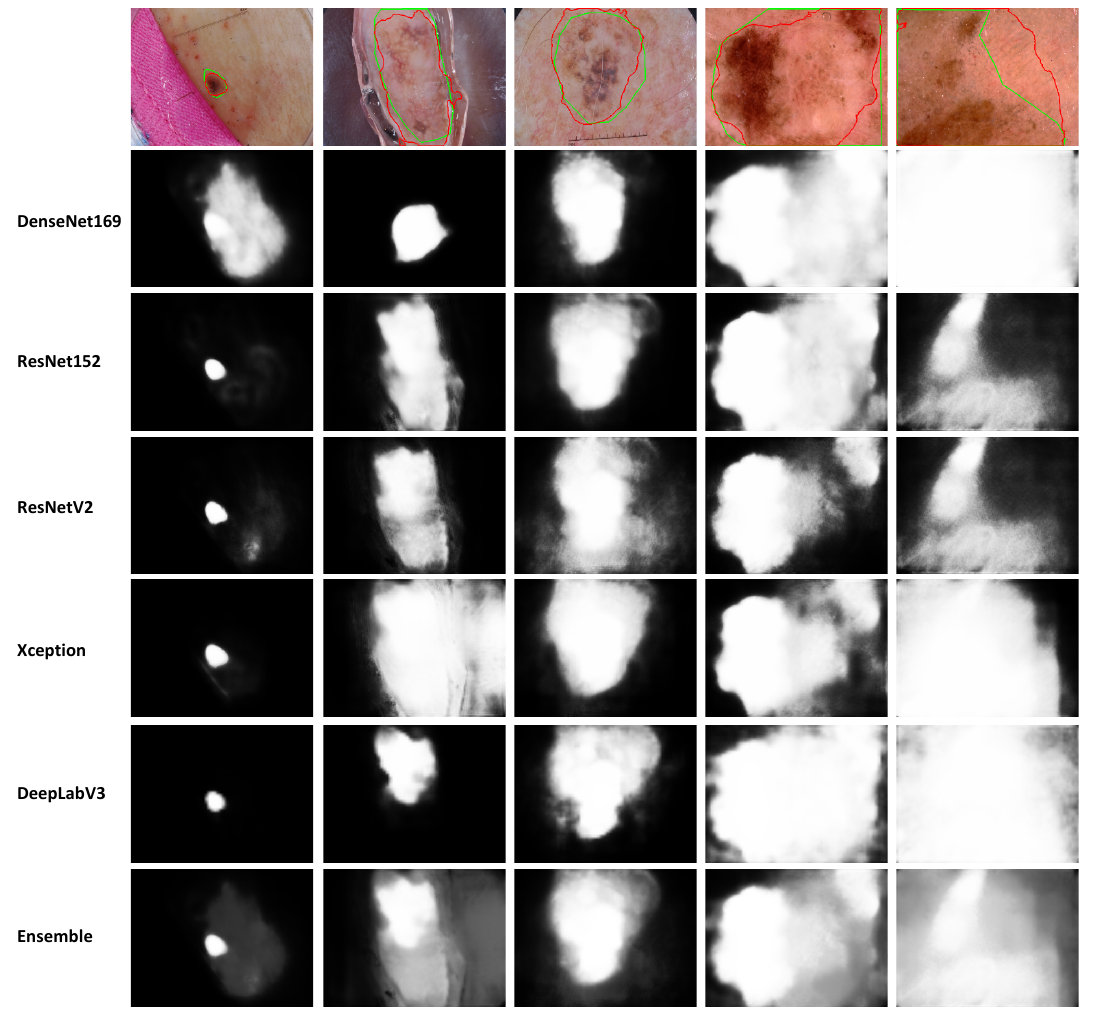}
	\caption{Prediction map of lesion segmentation framework using different base models. In the first row, green contours represent the ground truth and red contours delineate proposed segmentation output (achieved from thresholding the ensembling of predictions). The second to the last rows depict prediction maps generated by different base models and their ensemble.}
\label{fSeg}
\end{figure*}
\subsubsection{ISIC2017}
For the ISIC2017 data set, our segmentation networks (with different base models) are trained also in a 5-fold experiment framework for each model and best thresholding parameters are selected based on their outcomes. Evaluation results for ISIC2017 test obtained from different base models and their ensemble output are reported in the Table \ref{tISIC2017_segTest} alongside with results from other state-of-the-art methods. Methods in the Table \ref{tISIC2017_segTest} are sorted based on JSI criterion, as this metric was originally used to rank the submissions in ISIC2017 challenge. The last six rows in the Table \ref{tISIC2017_segTest} refer to our proposed methods.

As you can see in the Table \ref{tISIC2017_segTest}, for ISIC2017 Task 1 test set, even some of our single model predictions outperformed the most powerful methods the literature based on JSI metric. Four of our segmentation networks (with DenseNet169, ResNet1592, Xception, and ResNetV2 base models) are able to outperform the most powerful method reported in the literature, Chen et al. \cite{chen2018} by a high margin. The least and the most powerful methods of ours on the ISIC2017 test set, DenseNet169 and ResNetV2, performed 0.3\% and 1.5\% better than Chen et al method \cite{chen2018} in terms of JSI metric, respectively. This transcendence of our proposed methods against the best method from the literature is exceeded to 1.9\% for the ensemble prediction, which set a new state-of-the-art result for the ISIC2017 Task 1 test data set. It is important to remark that other evaluation metrics (ACC, DSC, and SEN) reported for our proposed method are higher than any other reported method in the survey.

\begin{table}[]
\centering
\caption{Evaluation metrics reported for ISIC2017 Task1 test set resulted from different lesion segmentation methods}
\begin{threeparttable}
  \centering
\begin{tabular}{llllll}
\hline \hline
\multirow{2}{*}{\textbf{Methods}\tnote{$\ddagger$} } & \multicolumn{5}{l}{\textbf{Averaged Evaluation Metrics (\%)}}            \\ \cline{2-6} 
                                  & \textbf{ACC} & \textbf{DSC} & \textbf{JSI} & \textbf{SEN} & \textbf{SPC} \\ \hline
Unet \cite{UNET}                          & 92.0         & 76.8         & 65.1         & 85.3         & 95.7         \\
Garcia et al. \cite{jose2019}                          & 88.4         & 76.0         & 66.5         & 86.9         & 92.3         \\
FCN-8  \cite{FCN}                         & 93.3         & 78.3         & 69.6         & 80.6         & 95.4         \\
Galdran et al.(*8)   \cite{galdran2017}                & 92.3         & 82.4         & 73.5         & 81.3         & 96.8         \\
Jahanifar et al. (*7)  \cite{jahanifar2018supervised}               & 93.0         & 83.9         & 74.9         & 81.0         & 98.1         \\
Kawahara et al. (*6)    \cite{kawahara2018fully}              & 93.0         & 83.7         & 75.2         & 81.3         & 97.6         \\
Yuexiang Li (*9) \cite{li2018}              & 95.0         & 83.9         & 75.3         & 85.5         & 97.4         \\
Menegola et al. (*5)    \cite{menegola2017}              & 93.7         & 83.9         & 75.4         & 81.7         & 97.0         \\
Lei Bi et al. (*3,*4)   \cite{bi2017ISIC}             & 93.4         & 84.4         & 76.0         & 80.2         & 98.5         \\
Xinzi He et al.          \cite{he2018}                  & 93.9         & 84.5         & 76.1         & -            & -            \\
Berseth (*2) \cite{berseth2017}                  & 93.2         & 84.7         & 76.2         & 82.0         & 97.8         \\
Yuan et al.(*1)    \cite{yuan2017}                  & 93.4         & 84.9         & 76.5         & 82.5         & 97.5         \\
Hang Li et al.    \cite{li2018JBHI}                  & 93.9         & 86.6         & 76.5         & 82.5            & 98.4            \\
Navarro et al.     \cite{navarro2018}                  & 95.5         & 85.4         & 76.9         & -            & -            \\
Tschandl et al.    \cite{tschandl2019}                  & -            & 85.3         & 77.0         & -            & -            \\
Al-masani et al.   \cite{almasani2018}                  & 94.0         & 87.0         & 77.1         & 85.4         & 96.6         \\
Lei Bi et al.         \cite{bi2019}                  & 94.0         & 85.6         & 77.7         & 86.2         & 96.7         \\
Sarker et al. \cite{sarker2018}               & 93.6         & 87.8         & 78.5         & -            & -            \\
Xue et al.            \cite{xue2018}              & 94.1         & 86.7         & 78.5         & -            & -            \\
Sheng Chen et al.        \cite{chen2018}                  & 94.4         & 86.8         & 78.7         & -            & -            \\ 
Xiaomeng Li et al. \cite{xiaomeng2018}                           & 94.3         & 87.4         & 79.8         & 87.9         & 95.3         \\ \hline
\textit{deeplabv3}                & 94.3         & 85.8         & 77.8         & 83.5         & 98.2         \\
\textit{DenseNet169}              & 93.8         & 86.8         & 79.0         & 86.8         & 96.3         \\
\textit{ResNet152}                & 94.1         & 87.1         & 79.4         & 87.7         & 96.1         \\
\textit{Xception}                 & 94.4         & 87.4         & 79.8         & 86.4         & 97.1         \\
\textit{ResNetV2}                 & 94.4         & 87.6         & 80.2         & 87.3         & 96.4         \\
\textit{Ensemble}                 & 94.6         & 87.9         & 80.6         & 87.9         & 96.9         \\ \hline \hline
\end{tabular}
\begin{tablenotes}
    \item [$\ddagger$] The *-marked numbers in the parentheses shows the rank of that method in the original ISIC2017 Task 1 challenge leaderboard.% (https://challenge.kitware.com/\#phase/584b0afacad3a51cc66c8e24).
    \end{tablenotes}
\end{threeparttable}%
\label{tISIC2017_segTest}
\end{table}

\subsubsection{ISIC2016}
We evaluate our proposed lesion segmentation method on ISIC2016 test set like the way we did for ISIC2017 data set. The resulted metrics on ISIC2016 test set are reported in Table \ref{tISIC2016_segTest}, in which results from the other method in the literature and ISIC2016 original leaderboard submissions are reported too. Regarding \ref{tISIC2016_segTest} and based on JSI evaluation metric, from our proposed segmentation models, the Xception and ResNetV2 based segmentation networks are able to perform the same and even better than He et al. method \cite{he2018}. Considering the results from \cite{he2018} as highest reported metrics in the literature, ResNetV2 based outperform it with a marginal difference of 0.1\% and setting a new state-of-the-art for ISIC2016 data set. The ensemble prediction of all base networks, however, is ranked first with a larger margin based on the JSI criterion. High values other evaluation metrics (ACC, DSC, SEN, SPC) reported for our models, approves the excellence of our proposed skin lesion segmentation framework among other reviewed methods. 

\begin{table}[]
\centering
\caption{Evaluation metrics reported for ISIC2016 Task1 test set resulted from different lesion segmentation methods}
\begin{threeparttable}%
\centering
\begin{tabular}{llllll}
\hline \hline
\multirow{2}{*}{\textbf{Method}\tnote{$\top$}}    & \multicolumn{5}{l}{\textbf{Averaged Evaluation Metrics (\%)}} \\ \cline{2-6} 
                           & \textbf{ACC}      & \textbf{DSC}      & \textbf{JSI}      & \textbf{SEN}      & \textbf{SPC}      \\ \hline
%burdick2017                & 69.3     & -        & -        & 76.0     & 72.3     \\
Fan et al. \cite{fan2017}                & 89.3     & 79.2     & -        & 73.8     & 94.0     \\
Ahn et al. \cite{ahn2017}                    & -        & 83.4     & -        & -        & -        \\
Unet    \cite{UNET}                   & 93.6     & 86.8     & 78.2     & 93.0     & 93.5     \\
Garcia et al.    \cite{jose2019}               & 93.4     & 86.9     & 79.1     & 87.0     & 97.8     \\
Zamani et al. (*5)     \cite{tajeddin2016general}            & 94.6     & 88.8     & 81.0     & 83.2     & 98.7     \\
Jeremy Kawahara$\star$ (*4)                 & 94.4     & 88.5     & 81.1     & 91.5     & 95.5     \\
FCN-8    \cite{FCN}                   & 94.1     & 88.6     & 81.3     & 91.7     & 94.9     \\
Mahmudur Rahman$\star$ (*3)                    & 95.2     & 89.5     & 82.2     & 88.0     & 96.9     \\
Yu et al. (*2)  \cite{yu2017TMI}             & 94.9     & 89.7     & 82.9     & 91.1     & 95.7     \\
Mirikharaji et al. \cite{mirikharaji2018}                & 95.0     & 90.1     & 83.3     & 90.1     & 96.6     \\
Lei Bi et al. \cite{bi2017ISBI}              & 95.1     & 90.2     & 83.3     & 91.8     & 95.2     \\
Jahanifar et al.  \cite{jahanifar2018supervised}            & 94.3     & 90.7     & 83.8     & 90.1     & 96.6     \\
Urko Sanchez$\star$ (*1)                    & 95.3     & 91.0     & 84.3     & 91.0     & 96.5     \\
Codelalla et al. \cite{codella2017skin}             & 95.3     & -        & 84.3     & -        & -        \\
Lei Bi et al.  \cite{bi2017TBME}             & 95.5     & 91.1     & 84.6     & 92.1     & 96.5     \\
Yuan et al. \cite{yuan2017TMI}             & 95.5     & 91.2     & 84.7     & 91.8     & 96.6     \\
Yuan et al.    \cite{yuan2017}               & 95.7     & 91.3     & 84.9     & 92.4     & 96.5     \\
Lei Bi et al.  \cite{bi2019}                   & 95.8     & 91.7     & 85.9     & 93.1     & 96.0     \\
Hang Li et al.   \cite{li2018JBHI}                  & 95.9     & 93.1     & 87.0     & 95.1        & 96.0        \\
Xinzi He et al.  \cite{he2018}                   & 96.0     & 93.1     & 87.1     & -        & -        \\ \hline
\textit{ResNet152}         & 95.9     & 92.3     & 86.2     & 93.6     & 96.0     \\
\textit{DenseNet169}       & 96.1     & 92.6     & 86.6     & 92.2     & 96.9     \\
\textit{deeplabv3}         & 96.2     & 93.0     & 87.0     & 93.6     & 97.0     \\
\textit{Xception}          & 96.2     & 93.2     & 87.1     & 93.6     & 96.9     \\
\textit{ResnetV2}          & 96.3     & 93.3     & 87.2     & 93.8     & 97.1     \\
\textit{\textbf{Ensemble}} & 96.5     & 93.4     & 87.4     & 94.5     & 96.6     \\ \hline \hline
\end{tabular}
\begin{tablenotes}
    \item [$\top$] No cite-able manuscript has been found for methods marked with $\star$ symbol.  The *-marked numbers in the parentheses shows the rank of that method in the original ISIC2016 Task 1 challenge leaderboard. %(https://challenge.kitware.com/\#phase/566744dccad3a56fac786787).
    
    \end{tablenotes}
\end{threeparttable}%
\label{tISIC2016_segTest}
\end{table}

\subsubsection{Qualitative results}
To show the power of the proposed segmentation method and the positive effect of model ensembling, outputs of segmentation models and the final segmentation mask are illustrated for five samples in Fig. \ref{fSeg}. 

In the first row of Fig. \ref{fSeg} original images are depicted with the boundaries of lesion overlaid on them. In that images, green contours represent ground truth boundaries, and red contours show the boundaries of the proposed segmentation framework. The second to seventh rows of Fig. \ref{fSeg} illustrate the output predictions of segmentation framework with DensNet169, ResNet152, ResNetV2, Xception, and DeepLab, respectively. The final row in Fig. \ref{fSeg} is the ensemble prediction obtained by averaging all the above predictions.

As illustrated in Fig. \ref{fSeg}, DenseNet169 shows a good performance on three samples (column 3rd to 6th)  but it does not perform well on the first two columns (left to right), whereas other networks performed well enough on those two samples. Therefore, by benefiting from the ensembling, we could achieve a plausible result in the last row of Fig. \ref{fSeg}. The same phenomenon happened almost in all samples, where ensemble prediction outperformed each individual model predictions.

%%%%%%%%%%%%%---------------------------------------------

\subsection{Attributes segmentation results}
\subsubsection{ISIC2018}
As inferred from Tables \ref{tAttTrain}, for each attribute detection task the segmentation framework with DenseNet169 base models outperforms other single-model predictions, except for “Milia like cysts” attribute for which ResNetV2 performs slightly better. This fact also holds for “All Attributes” situation  where DenseNet169 gains C-DSC of 62.9\% and C-JSI of 46.4\%. As expected, ensemble predictions elevated the results for every single model for every attribute by a high margin, achieving C-DSC of 65.1\% and C-JSI of 48.3\% for the “All Attribute” case which is about 2\% better than the best performing single-model (DenseNet169).

\begin{table*}[]
\centering
\caption{ Evaluation metrics obtained from cross-validation experiments on ISIC2018 Task2 training set using different base models in the proposed attribute detection framework.}
\resizebox{.95\textwidth}{!}{%
\begin{tabular}{lcc|cc|cc|cc|cc|cc}
\hline \hline
\multirow{2}{*}{\textbf{Base Network}} & \multicolumn{2}{c|}{\textbf{Pigment Networks}} & \multicolumn{2}{c|}{\textbf{Globules}} & \multicolumn{2}{c|}{\textbf{Milia like Cysts}} & \multicolumn{2}{c|}{\textbf{Negative Networks}} & \multicolumn{2}{c|}{\textbf{Streaks}} & \multicolumn{2}{c}{\textbf{All Atributes}} \\ \cline{2-13} 
                              & C-JSI            & C-DSC             & C-JSI        & C-DSC         & C-JSI            & C-DSC             & C-JSI             & C-DSC             & C-JSI        & C-DSC        & C-JSI        & C-DSC        \\ \hline
ResNet151                     & 0.527              & 0.690            & 0.304          & 0.466        & 0.144              & 0.257            & 0.149               & 0.260            & 0.125          & 0.222       & 0.436          & 0.598       \\ 
Resnetv2                      & 0.539              & 0.706            & 0.310          & 0.473        & 0.159              & 0.274            & 0.189               & 0.318            & 0.121          & 0.216       & 0.455          & 0.616       \\ 
DenseNet169                   & 0.538              & 0.699            & 0.324          & 0.490        & 0.158              & 0.273            & 0.213               & 0.351            & 0.134          & 0.236       & 0.464          & 0.629       \\ 
Ensemble                      & 0.563              & 0.720            & 0.341          & 0.508        & 0.171              & 0.289            & 0.228               & 0.371            & 0.156          & 0.270       & 0.483          & 0.651       \\ 
%ji2018_attributes             & 0.382              & -                & 0.278          & -            & 0.128              & -                & 0.227               & -                & 0.254          & -           & -              & -           \\ \hline
\hline \hline
\end{tabular}
}
\label{tAttTrain}
\end{table*}

\begin{table*}[ht]
\centering
\caption{Evaluation metrics resulted from ISIC2018 Task 2, attribute detection, testing set}
\begin{threeparttable}
  \centering
    \begin{tabular}{lllll}
    \hline \hline
    \multirow{2}{*}{\textbf{Rank}} & \multirow{2}{*}{\textbf{User}} & \multirow{2}{*}{\textbf{Method Description}\tnote{$\bot$}}                                          & \multicolumn{2}{l}{Evaluation Metrics} \\ \cline{4-5} 
                          &                       &                                                                              & \textbf{C-DSC}              & \textbf{C-JSI}             \\ \hline
    1                     & Navid Alemi           & Proposed method-Ensembled prediction-with lesion mask                        & 64.2               & 43.4              \\
    2                     & Navid Alemi           & Proposed method-Ensembled prediction-without lesin mask                      & 64.0               & 47.1              \\
    3                     & Navid Alemi           & Proposed method-Ensembled prediction-0.05 threshold decreasement              & 63.9               & 47.0              \\
    4                     & Lenovo E-Health       & Multi-level feaure fusion in Pyramid Scene Parsing Network      & 60.6               & 43.4              \\
    5                     & Jeffery Wu            & Unet without validation \cite{chen2018attribute}                           & 60.5               & 43.3              \\
    6                     & Jeffery Wu            & Unet with validation \cite{chen2018attribute}                              & 60.3               & 43.2              \\
    7                     & Lenovo E-Health       & Pyramid Scene Parsing Network with sigmoid activation function & 59.8               & 42.7              \\
    8                     & Lijun Gong            & Automatic Skin Lesion Attribute Detection with Deep Networks\tnote{*1}             & 58.4               & 41.2              \\
    9                     & Xu Min                & Three step Unet with majarity voting ensemble \tnote{*2}                           & 57.0               & 39.9              \\
    10                    & Xu Min                & Three step Unet with weighted average ensemble \tnote{*2}                          & 57.0               & 39.9              \\
    11                    & Xu Min                & Three step Unet with unweghted average from 24 selected best models\tnote{*2}      & 56.9               & 39.8              \\
    12                    & Yuming Qiu            & Automated Visual‐attention‐guided Approach with ROI‐unifying DCNNs \tnote{*3}      & 55.0               & 37.9              \\
    13                    & Andrey Sorokin        & Mask-RCNN with SGD \tnote{*4}                                                      & 54.4               & 37.4              \\
    14                    & RECOD Titans          & Average of 4 Deep-Learning Models with Thresholding \cite{bissoto2018}        & 51.1               & 34.4              \\ \hline
    \end{tabular}
    \begin{tablenotes}
    \item [$\bot$] Results in this table are directly adopted from the ISIC2018 challenge Task 2 test leaderboard. %(https://challenge.kitware.com/\#phase/5b1c193356357d41064da2ec). 
    Entries marked with "*" symbol were not accompanied by a cite-able article. However, their submission report titles are listed below:
        \item[*1] Automatic Skin Lesion Analysis with Deep Networks
        \item[*2] U-Net Ensemble for Skin Lesion Analysis towards Melanoma Detection
        \item[*3] Methods Description for ISIC2018 Challenge
        \item[*4] Lesion Analysis and Diagnosis with Mask-RCN
    \end{tablenotes}
\end{threeparttable}%
\label{tISIC2018_attTest}
\end{table*}

Best performing models and parameters based on cross-validation experiments were selected to predict the segmentation maps for ISIC2018 Task 2 testing set, which contain 1000 dermoscopic images, and their results were uploaded to the ISIC2018 Task 2 challenge host in order to be evaluated. Evaluation result in this category are directly adopted from the challenge leader-board and reported in Table \ref{tISIC2018_attTest}. Our best performing results attained C-DSC of 64.2\% and C-JSI of 47.3\% which brought us on top in the ISIC2018 Task 1 challenge leader-board. The large gap between our best submission and the next best-performing competitor (about 4\% improvement in both C-JSI and C-DSC metrics) approves the distinction power of our proposed attribute segmentation method over other state-of-the-art algorithms. Moreover, gaining higher evaluation metrics for the test set compare to the train set implying the strong generalizability of the proposed method. It worth mentioning that we have three different submissions for ISIC2018 Task 2 challenge and all of them performed considerably better than other competitors. The first submission used all networks for all attributes and applied the lesion mask restriction on its prediction in the post-processing tasks. The second submission used all networks but without applying lesion mask restriction. The third submission was similar to the first submission (with skin lesion restriction applied), but the ${T_H}$ threshold that was applied for all attributes in that submission is decreased  by value of 0.05  to keep more objects in the ${Attribute_{marker}}$.%

\subsubsection{Qualitative results}
\begin{figure*}[t]
	\centering
    \includegraphics[width=0.90\textwidth]{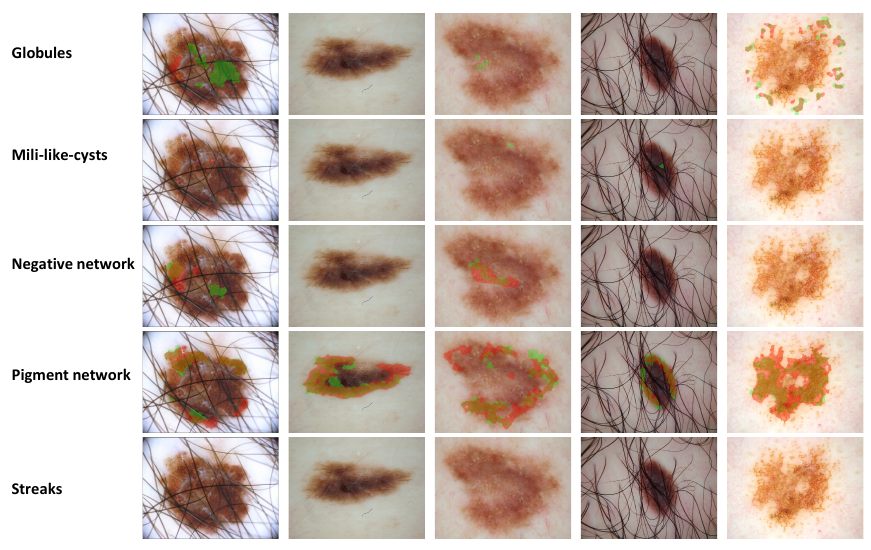}
	\caption{Outputs of proposed attribute segmentation framework. Each row represents a specific attribute annotation. Ground truth regions, the predicted regions by our framework, and intersections of regions are highlighted with green, red and orange, respectively.}
	\label{fAtt}
\end{figure*}

Output segmentations of different attributes for five sample dermoscopic images are illustrated in Fig. \ref{fAtt}, in which, each column shows a sample and each row is associated with one specific attribute. Order of attributes from the first row to the last row in Fig. \ref{fAtt} is as follows: globules, Milia-like cysts, negative network, pigment network, and streaks. In each image of Fig. \ref{fAtt}, ground truth of attributes are overlaid with green color on the image, proposed segmentations are depicted with red color, and their intersection is specified by orange color.

The proposed attribute segmentation algorithm is able to successfully predict regions with different attribute labels (intersection regions are rather large). Even for cases with extreme hair, like the first column in Fig. \ref{fAtt}, existing attributes are clearly detected. 
%%% MUST ADD MORE COMMENTS WHEN IMAGES ARE REPLACED

\section{Discussion} \label{sec:discussion}
In this paper, we proposed two main contributions to the skin lesion and their attributes segmentation tasks. These contributions are 1) a transfer learning based segmentation architecture which includes powerful pre-trained deep neural networks in its encoding path, skip connections, multi-scale convolutional blocks and pyramid pooling paradigm in the decoding path and 2) a set of novel, practical, and domain-specific data augmentation techniques. The prediction resulted from our proposed segmentation network are then post-processed to form the final segmentation of the lesions or their attributes. As reported in section \ref{sec:setups}, our segmentation framework was able to outperform other state-of-the-art approaches with a considerable margin. However in this section, we gain insight into the effects that each contribution makes to the initial predictions i.e., the output of the segmentation network without any post-processing.

To be able to assess the raw predictions of the network, we incorporate Receiver Operating Characteristics (ROC) curves and Area Under Curve (AUC) criteria. Higher values of AUC shows better performance of the network in predicting  labels. To draw ROC curves, we must consider segmentation as a pixel classification task in which our network works as a binary classifier trying to decide whether a pixel belongs to positive or negative classes. ROC curve is then created by plotting classifier sensitivity (True Positive Rate: TPR) against classification fall-out (False Positive Rate: FPR) at different threshold settings \cite{hajian2013ROC}.
%The AUC criterion is the area under the ROC curve. Based on the ROC curve axis limits, minimum and maximum values for AUC would be 0 and 1. Higher AUC values imply a better classification performance. The best ROC curve is the one which its extents is nearest to the left vertical and top horizontal axes, this situation would lead to higher AUC values. In other words, the higher the ROC curve, the higher the AUC will be.

\begin{figure*}[t]
	\centering
    \includegraphics[width=0.8\textwidth]{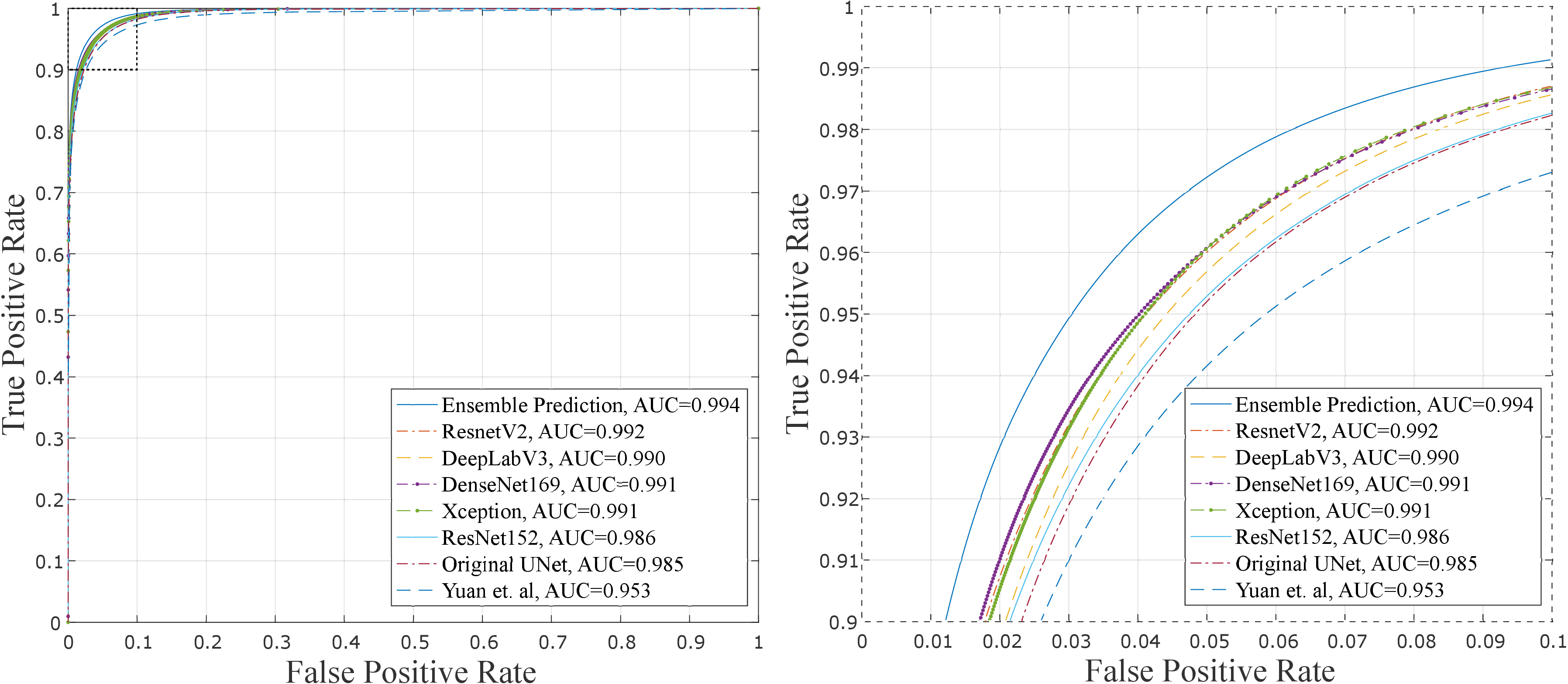}
	\caption{Left: ROC curves of lesion predictions obtained form cross-validation experiments on ISIC2018 Part 1 training data set using different base models or network architectures. Right: Zoomed view of the selected area in the left figure. AUC values for each method is reported in the legend.}
	\label{fig:roc_lesion}
\end{figure*}

Here, we assess the effect of each major contribution on the performance of the proposed framework for lesion/attribute detection. To this end, output predictions resulted from cross-validation experiments on ISIC2018 training data set were  used. First, to understand the performance quality of our proposed transfer learning based segmentation network against other conventional U-Net like networks, we used lesion detection as the reference task. Fig. \ref{fig:roc_lesion} shows ROC curves that are created based on the output predictions achieved using our transfer learning framework with different base networks (densenet169, resnetv2, deeplabv3, xception, and resnet152). ROC curves originated from outputs of original U-Net and Yuan et al. \cite{yuan2017} networks are also depicted. As one can see in Fig. \ref{fig:roc_lesion}, no matter which base network have been used, our transfer learning based network was able to outperform the conventional U-Net and ISIC2017 Part1 (Lesion Segmentation Challenge) winner, Yuan et al. \cite{yuan2017} approach. Higher AUC scores from transfer learning based networks, as well as upper ROC curves, approve the superiority of our proposed segmentation networks. 
Another observation that can be revealed by Fig. \ref{fig:roc_lesion} is performance of the ensemble prediction with an AUC value of 0.994 is much better than any other single network performance, which approves previous claims about the power of prediction ensembling. 
Difference between AUC values is very minimal that represents transfer learning approaches with different base models are performing as good as each other. However, based on ROC curve and AUC values reported in Fig.  \ref{fig:roc_lesion}, ResNetV2 base model gives the best performing single network prediction for lesion segmentation, which is correlated with the final segmentation results in Table \ref{tSegTrain}. Please note that we used the same experiment setups to obtain all the curves in Fig. \ref{fig:roc_lesion}. 

In dense prediction tasks (like image segmentation), small increment in ROC and AUC would make a considerable improvement in the final segmentation output. Based on Table \ref{tSegTrain}, the best results from ensemble prediction achieves a value of 0.8142 for Thresholded Jaccard (TJI) criterion and deeplabv3 results to TJI of 0.7932, which indicates a considerable difference of 2\% in final segmentation performance. However, assessment of raw predictions from these two methods in terms of AUC criterion shows only 0.4\% difference based on values reported in Fig. \ref{fig:roc_lesion}. This intuitive comparison shows that small improvements in network predictions (based on ROC and AUC) leads to a considerable positive effect in the final segmentation output, and yet, ROC and AUC criterions are suitable enough to assess the network performance without taking any post-processing steps.
Usually, the lesion is a relatively big object in the image that covers a considerable portion of image pixels. Therefore, detecting a lesion area is a simpler task in comparison to attributes detection. This is also the main reason we have high values reported for AUC criterion in the lesion segmentation task.

\begin{figure}[t]
	%\centering
    \includegraphics[width=0.8\columnwidth]{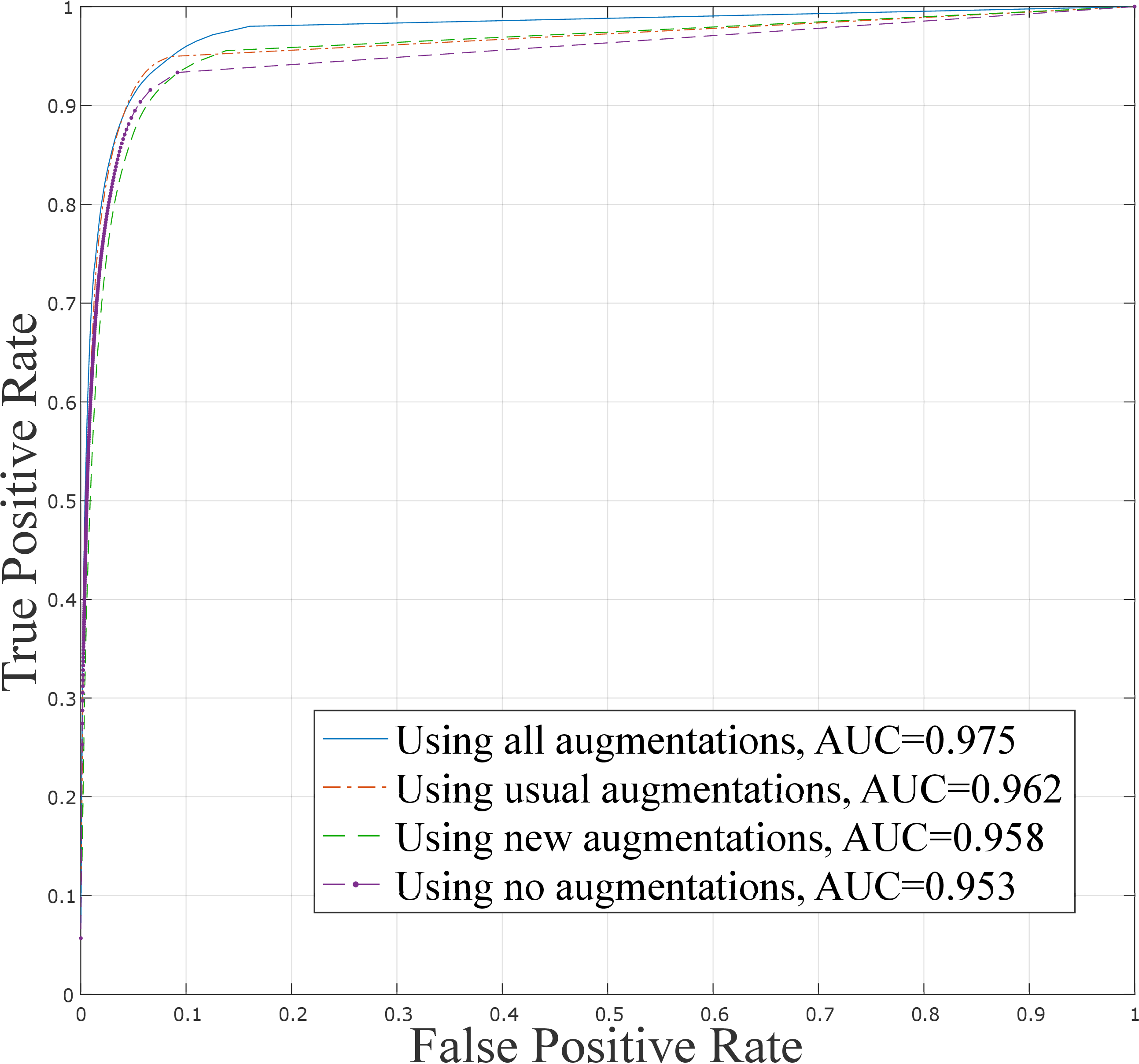}
    \centering
	\caption{ROC curves of "Negative Networks" attribute predictions obtained form cross-validation experiments on ISIC2018 Part 2 training data set using proposed transfer learning based network with "densenet169" base model. Each curve correspond to different data augmentation scenario.}
	\label{fig:roc_attribute}
\end{figure}

To investigate the effect of our data augmentation techniques in the predictions of our proposed network, we would assess the performance of our transfer learning architecture with the “densenet169” as its base network for attribute detection task under different data augmentation paradigms. In this regard, four different data augmentation scenarios have been investigated in the “Negative Network” attribute detection task. In other words, we would have four transfer learning setups with same base network and similar training configuration but different data augmentation techniques, each performing 5-fold cross-validation experiments on ISIC2018-Part2 training data set. 
Data augmentation scenarios used in these experiments are: 1) using no data augmentation techniques, 2) using only conventional data augmentation techniques (horizontal/vertical flip, translation, zooming, shearing, color jittering, intensity scaling, rotation, and adding noise), 3) using only introduced augmentation techniques (contrast adjustment, sharpness adjustment, disturbing illumination, and hair occlusion), and 4) using all available data augmentation techniques. The ouput of incorporating each of these data augmentation scenarios is evaluated by  ROC curve and AUC criterion, and the results are illustrated in Fig. \ref{fig:roc_attribute}. 

As it is obvious from Fig. \ref{fig:roc_attribute}, using all augmentation techniques can outperform all other scenarios by a large margin, achieving AUC value of 0.975 in comparison to the state with no data augmentation which achieves AUC of 0.953. This high difference in performance shows the importance of using data augmentation in attribute detection task. However, using only usual augmentation techniques would lead to AUC of 0.962 which is only 0.004 better than using only newly introduced augmentation techniques. However, it is critical to mention that introduced data augmentation techniques are able to boost the usual data augmentation techniques performance reach their best performance (AUC of 0.975 which is 1.3\% improvement in performance). These inferences can be easily drawn from ROC curves, qualitatively, since they are showing a similar behavior based on Fig. \ref{fig:roc_attribute} (ROC curve related to using all data augmentation techniques stands above all other curves). 
Fig. \ref{fig:roc_attribute} proves that our proposed augmentation techniques are able to elevate the performance of densenet169 on detecting Negative Network attribute. Similar results can be achieved when experimenting with other base networks or attributes. We chose Negative Network attribute because it is relatively hard and rare to detect. The Negative Networks are very thin hair-like structures with low contrast appearance against the lesion texture. They are usually lighter than the dominant color of the lesion and can be easily missed as white-colored hairs in the image. Using proposed augmentation techniques like adding hair occlusion, reducing the contrast and sharpness in the training phase can make the segmentation network robust against usual deficiencies in the image. Moreover, increasing the contrast and sharpness during the test time (predicting the samples mask using trained networks), would help to elevate the results. In the same manner, detection of streaks and pigment network attributes will be enhanced using introduced augmentation techniques as they are thin and hair-like structures with weak appearance in the image. These three novel augmentation techniques (hair occlusion, sharpness and contrast adjustment) would be very practical and useful when detection of small objects like dot and globules is our objective. However, disturbing the illumination gradient is beneficial during dominant object detection tasks like lesion segmentation. Especially, for cases that lesion extent reaches close to the image boundary and due to uneven illumination and dark shadows in those areas, lesion area would be faded into the darkness and the border between the lesion and normal skin becomes less distinguishable. By disturbing illumination during the network training phase, we can make the network more robust against this bad effect.

One may argue that incorporating various network architectures would cause network training taking too long, but, in practice, this was not a challenge. That is because our transfer learning based segmentation network, which benefits from pre-trained model weights, is able to converge in less than 70 epochs and achieve their best performances. In comparison with a state-of-the-art network \cite{yuan2017TMI} which was trained from scratch till 600 epochs, our five networks all together need no more than 350 epochs, totally. Using transfer learning also reduces the risk of over-fitting and enhance the generalizability of our segmentation network in the prediction of unseen data.

Future direction for this work is to leverage transfer learning with the domain specific knowledge. Like the way Tschandl et al. and Bi et al. proposed in their recent works \cite{tschandl2019,bi2017ISIC,bi2019}, it is a good practice to fine-tune the base models of our proposed segmentation network (encoding path) on a bigger data set of dermoscopic images, instead of using ImageNet pre-trained network weights that were trained on natural images. Fortunately, there are big data sets available for lesion recognition task, like \cite{Tschandl2018_HAM10000} which contain over 10000 dermoscopic images of common skin lesions in different classes. Training and fine-tuning our base models on this data set (as a lesion classifier) will make it able to extract more relevant and discriminant feature maps to be fed into the decoder path of our proposed segmentation network.

\section{Conclusion} \label{sec:conclusion}
In this paper, we proposed to use transfer learning and extensive ensemble of predictions in order to boost performance on the skin lesion and its attributes segmentation task. Proposed segmentation framework incorporates a  pre-trained base networks in its encoding path, and multi-scale convolutional and pyramid pooling blocks in the decoding path. Furthermore, several on-fly data augmentation techniques were utilized, some of which were specially tailored to mimic the variation present in the dermoscopic images. Our models have been evaluated in a 5-fold cross-validation framework to reduce the risk of over-fitting and generating a basis for optimal hyper-parameters estimation. Evaluation experiments on the training and validation set shows promising performance on both lesion boundaries segmentation and lesion attribute detection tasks, setting state-of-the-art results for all ISIC2016, ISIC2017, and ISIC2018 data sets.
\bibliographystyle{ieeetr}%{plainnat-fa}%{unsrt-fa}%{chicago-fa}{plainnat-fa}%

% \bibliographystyle{ieeetr}%{plainnat-fa}%{unsrt-fa}%{chicago-fa}{plainnat-fa}%
% \bibliography{MyCollection.bib}

\begin{thebibliography}{10}

\bibitem{oliveira2016computational}
R.~B. Oliveira, E.~Mercedes~Filho, Z.~Ma, J.~P. Papa, A.~S. Pereira, and
  J.~M.~R. Tavares, ``Computational methods for the image segmentation of
  pigmented skin lesions: a review,'' {\em Computer methods and programs in
  biomedicine}, vol.~131, pp.~127--141, 2016.

\bibitem{gutman2016skin}
D.~Gutman, N.~C. Codella, E.~Celebi, B.~Helba, M.~Marchetti, N.~Mishra, and
  A.~Halpern, ``Skin lesion analysis toward melanoma detection: A challenge at
  the international symposium on biomedical imaging (isbi) 2016, hosted by the
  international skin imaging collaboration (isic),'' {\em arXiv preprint
  arXiv:1605.01397}, 2016.

\bibitem{codella2017skin}
N.~C. Codella, D.~Gutman, M.~Emre~Celebi, B.~Helba, M.~A. Marchetti, S.~W.
  Dusza, A.~Kalloo, K.~Liopyris, N.~Mishra, H.~Kittler, and A.~Halpern, ``Skin
  lesion analysis toward melanoma detection: A challenge at the 2017
  international symposium on biomedical imaging (isbi), hosted by the
  international skin imaging collaboration (isic),'' {\em arXiv preprint
  arXiv:1710.05006}, 2017.

\bibitem{tajeddin2018melanoma}
N.~Zamani~Tajeddin and B.~Mohammadzadeh~Asl, ``Melanoma recognition in
  dermoscopy images using lesion's peripheral region information,'' {\em
  Computer Methods and Programs in Biomedicine}, vol.~163, pp.~143 -- 153,
  2018.

\bibitem{braun2019attributes}
K.~K. dermoscopedia – Ralph~Braun, ``Dermoscopic structures,'' 2019.
\newblock [Online; accessed 21-January-2019].

\bibitem{litjens2017survey}
G.~Litjens, T.~Kooi, B.~E. Bejnordi, A.~A.~A. Setio, F.~Ciompi, M.~Ghafoorian,
  J.~A. van~der Laak, B.~Van~Ginneken, and C.~I. S{\'a}nchez, ``A survey on
  deep learning in medical image analysis,'' {\em Medical image analysis},
  vol.~42, pp.~60--88, 2017.

\bibitem{yu2017TMI}
L.~Yu, H.~Chen, Q.~Dou, J.~Qin, and P.-A. Heng, ``Automated melanoma
  recognition in dermoscopy images via very deep residual networks,'' {\em IEEE
  transactions on medical imaging}, vol.~36, no.~4, pp.~994--1004, 2017.

\bibitem{diaz2018dermaknet}
I.~G. D{\'i}az, ``Dermaknet: Incorporating the knowledge of dermatologists to
  convolutional neural networks for skin lesion diagnosis,'' {\em accepted to
  be appeared in IEEE Journal of Biomedical and Health Informatics}, 2018.

\bibitem{Tschandl2018_HAM10000}
P.~Tschandl, C.~Rosendahl, and H.~Kittler, ``The {HAM10000} dataset, a large
  collection of multi-source dermatoscopic images of common pigmented skin
  lesions,'' {\em Scientific Data}, vol.~5, p.~180161, 2018.

\bibitem{Celebi2015}
M.~E. Celebi, Q.~Wen, H.~Iyatomi, K.~Shimizu, H.~Zhou, and G.~Schaefer, ``{A
  State-of-the-Art Survey on Lesion Border Detection in Dermoscopy Images},''
  in {\em Dermoscopy Image Analysis} (M.~E. Celebi, T.~Mendonca, and J.~S.
  Marques, eds.), pp.~97--129, CRC Press, 2015.

\bibitem{jahanifar2018supervised}
M.~Jahanifar, N.~Zamani~Tajeddin, B.~Mohammadzadeh~Asl, and A.~Gooya,
  ``Supervised saliency map driven segmentation of lesions in dermoscopic
  images,'' {\em accepted to be appeared in IEEE Journal of Biomedical and
  Health Informatics}, 2018.

\bibitem{fan2017}
H.~Fan, F.~Xie, Y.~Li, Z.~Jiang, and J.~Liu, ``Automatic segmentation of
  dermoscopy images using saliency combined with otsu threshold,'' {\em
  Computers in biology and medicine}, vol.~85, pp.~75--85, 2017.

\bibitem{ahn2017}
E.~Ahn, J.~Kim, L.~Bi, A.~Kumar, C.~Li, M.~Fulham, and D.~D. Feng,
  ``Saliency-based lesion segmentation via background detection in dermoscopic
  images,'' {\em IEEE journal of biomedical and health informatics}, vol.~21,
  no.~6, pp.~1685--1693, 2017.

\bibitem{jose2019}
J.~L. Garcia-Arroyo and B.~Garcia-Zapirain, ``Segmentation of skin lesions in
  dermoscopy images using fuzzy classification of pixels and histogram
  thresholding,'' {\em Computer methods and programs in biomedicine}, vol.~168,
  pp.~11--19, 2019.

\bibitem{tajeddin2016general}
N.~Zamani~Tajeddin and B.~Mohammadzadeh~Asl, ``A general algorithm for
  automatic lesion segmentation in dermoscopy images,'' in {\em Biomedical
  Engineering and 2016 1st International Iranian Conference on Biomedical
  Engineering (ICBME), 2016 23rd Iranian Conference on}, pp.~134--139, IEEE,
  2016.

\bibitem{guo2018}
Y.~Guo, A.~S. Ashour, and F.~Smarandache, ``A novel skin lesion detection
  approach using neutrosophic clustering and adaptive region growing in
  dermoscopy images,'' {\em Symmetry}, vol.~10, no.~4, p.~119, 2018.

\bibitem{navarro2018}
F.~Navarro, M.~Escudero-Vinolo, and J.~Bescos, ``Accurate segmentation and
  registration of skin lesion images to evaluate lesion change,'' {\em accepted
  to be appeared in IEEE Journal of Biomedical and Health Informatics}, 2018.

\bibitem{FCN}
J.~Long, E.~Shelhamer, and T.~Darrell, ``Fully convolutional networks for
  semantic segmentation,'' in {\em Proceedings of the IEEE conference on
  computer vision and pattern recognition}, pp.~3431--3440, 2015.

\bibitem{UNET}
O.~Ronneberger, P.~Fischer, and T.~Brox, ``U-net: Convolutional networks for
  biomedical image segmentation,'' in {\em International Conference on Medical
  image computing and computer-assisted intervention}, pp.~234--241, Springer,
  2015.

\bibitem{galdran2017}
A.~Galdran, A.~Alvarez-Gila, M.~I. Meyer, C.~L. Saratxaga, T.~Ara{\'u}jo,
  E.~Garrote, G.~Aresta, P.~Costa, A.~M. Mendon{\c{c}}a, and A.~Campilho,
  ``Data-driven color augmentation techniques for deep skin image analysis,''
  {\em arXiv preprint arXiv:1703.03702}, 2017.

\bibitem{berseth2017}
M.~Berseth, ``Isic 2017-skin lesion analysis towards melanoma detection,'' {\em
  arXiv preprint arXiv:1703.00523}, 2017.

\bibitem{xu2018}
H.~Xu and T.~H. Hwang, ``Automatic skin lesion segmentation using deep fully
  convolutional networks,'' {\em arXiv preprint arXiv:1807.06466}, 2018.

\bibitem{tschandl2019}
P.~Tschandl, C.~Sinz, and H.~Kittler, ``Domain-specific
  classification-pretrained fully convolutional network encoders for skin
  lesion segmentation,'' {\em Computers in biology and medicine}, vol.~104,
  pp.~111--116, 2019.

\bibitem{ResNet}
K.~He, X.~Zhang, S.~Ren, and J.~Sun, ``Deep residual learning for image
  recognition,'' in {\em Proceedings of the IEEE conference on computer vision
  and pattern recognition}, pp.~770--778, 2016.

\bibitem{chen2018}
S.~Chen, Z.~Wang, J.~Shi, B.~Liu, and N.~Yu, ``A multi-task framework with
  feature passing module for skin lesion classification and segmentation,'' in
  {\em Biomedical Imaging (ISBI 2018), 2018 IEEE 15th International Symposium
  on}, pp.~1126--1129, IEEE, 2018.

\bibitem{menegola2017}
A.~Menegola, J.~Tavares, M.~Fornaciali, L.~T. Li, S.~Avila, and E.~Valle,
  ``Recod titans at isic challenge 2017,'' {\em arXiv preprint
  arXiv:1703.04819}, 2017.

\bibitem{VGG}
K.~Simonyan and A.~Zisserman, ``Very deep convolutional networks for
  large-scale image recognition,'' {\em arXiv preprint arXiv:1409.1556}, 2014.

\bibitem{sarker2018}
M.~Sarker, M.~Kamal, H.~A. Rashwan, S.~F. Banu, A.~Saleh, V.~K. Singh, F.~U.
  Chowdhury, S.~Abdulwahab, S.~Romani, P.~Radeva, {\em et~al.}, ``Slsdeep: Skin
  lesion segmentation based on dilated residual and pyramid pooling networks,''
  {\em arXiv preprint arXiv:1805.10241}, 2018.

\bibitem{mirikharaji2018}
Z.~Mirikharaji, S.~Izadi, J.~Kawahara, and G.~Hamarneh, ``Deep auto-context
  fully convolutional neural network for skin lesion segmentation,'' in {\em
  Biomedical Imaging (ISBI 2018), 2018 IEEE 15th International Symposium on},
  pp.~877--880, IEEE, 2018.

\bibitem{xue2018}
Y.~Xue, T.~Xu, and X.~Huang, ``Adversarial learning with multi-scale loss for
  skin lesion segmentation,'' in {\em Biomedical Imaging (ISBI 2018), 2018 IEEE
  15th International Symposium on}, pp.~859--863, IEEE, 2018.

\bibitem{xiaomeng2018}
X.~Li, L.~Yu, H.~Chen, C.-W. Fu, and P.-A. Heng, ``Semi-supervised skin lesion
  segmentation via transformation consistent self-ensembling model,'' {\em
  arXiv preprint arXiv:1808.03887}, 2018.

\bibitem{almasani2018}
M.~A. Al-masni, M.~A. Al-antari, M.-T. Choi, S.-M. Han, and T.-S. Kim, ``Skin
  lesion segmentation in dermoscopy images via deep full resolution
  convolutional networks,'' {\em Computer methods and programs in biomedicine},
  vol.~162, pp.~221--231, 2018.

\bibitem{kawahara2018fully}
J.~Kawahara and G.~Hamarneh, ``Fully convolutional neural networks to detect
  clinical dermoscopic features,'' {\em accepted to be appeared in IEEE Journal
  of Biomedical and Health Informatics}, 2018.

\bibitem{li2018}
Y.~Li and L.~Shen, ``Skin lesion analysis towards melanoma detection using deep
  learning network,'' {\em Sensors}, vol.~18, no.~2, p.~556, 2018.

\bibitem{qian2018}
C.~Qian, T.~Liu, H.~Jiang, Z.~Wang, P.~Wang, M.~Guan, and B.~Sun, ``A detection
  and segmentation architecture for skin lesion segmentation on dermoscopy
  images,'' {\em arXiv preprint arXiv:1809.03917}, 2018.

\bibitem{he2017mask}
K.~He, G.~Gkioxari, P.~Doll{\'a}r, and R.~Girshick, ``Mask r-cnn,'' in {\em
  Computer Vision (ICCV), 2017 IEEE International Conference on},
  pp.~2980--2988, IEEE, 2017.

\bibitem{bi2017TBME}
L.~Bi, J.~Kim, E.~Ahn, A.~Kumar, M.~Fulham, and D.~Feng, ``Dermoscopic image
  segmentation via multi-stage fully convolutional networks,'' {\em IEEE Trans.
  Biomed. Eng}, vol.~64, no.~9, pp.~2065--2074, 2017.

\bibitem{bi2017ISIC}
L.~Bi, J.~Kim, E.~Ahn, and D.~Feng, ``Automatic skin lesion analysis using
  large-scale dermoscopy images and deep residual networks,'' {\em arXiv
  preprint arXiv:1703.04197}, 2017.

\bibitem{bi2019}
L.~Bi, J.~Kim, E.~Ahn, A.~Kumar, D.~Feng, and M.~Fulham, ``Step-wise
  integration of deep class-specific learning for dermoscopic image
  segmentation,'' {\em Pattern Recognition}, vol.~85, pp.~78--89, 2019.

\bibitem{yuan2017TMI}
Y.~Yuan, M.~Chao, and Y.-C. Lo, ``Automatic skin lesion segmentation using deep
  fully convolutional networks with jaccard distance,'' {\em IEEE Trans. Med.
  Imaging}, vol.~36, no.~9, pp.~1876--1886, 2017.

\bibitem{yuan2017}
Y.~Yuan and Y.-C. Lo, ``Improving dermoscopic image segmentation with enhanced
  convolutional-deconvolutional networks,'' {\em accepted to be appeared in
  IEEE Journal of Biomedical and Health Informatics}, 2017.

\bibitem{li2018JBHI}
H.~Li, X.~He, F.~Zhou, Z.~Yu, D.~Ni, S.~Chen, T.~Wang, and B.~Lei, ``Dense
  deconvolutional network for skin lesion segmentation,'' {\em accepted to be
  appeared in IEEE Journal of Biomedical and Health Informatics}, 2018.

\bibitem{he2018}
X.~He, Z.~Yu, T.~Wang, B.~Lei, and Y.~Shi, ``Dense deconvolution net: Multi
  path fusion and dense deconvolution for high resolution skin lesion
  segmentation,'' {\em Technology and Health Care}, vol.~26, no.~S1,
  pp.~307--316, 2018.

\bibitem{bissoto2018}
A.~Bissoto, F.~Perez, V.~Ribeiro, M.~Fornaciali, S.~Avila, and E.~Valle,
  ``Deep-learning ensembles for skin-lesion segmentation, analysis,
  classification: Recod titans at isic challenge 2018,'' {\em arXiv preprint
  arXiv:1808.08480}, 2018.

\bibitem{sorokin2018}
A.~Sorokin, ``Lesion analysis and diagnosis with mask-rcnn,'' {\em arXiv
  preprint arXiv:1807.05979}, 2018.

\bibitem{chen2018attribute}
E.~Z. Chen, X.~Dong, J.~Wu, H.~Jiang, X.~Li, and R.~Rong, ``Lesion attributes
  segmentation for melanoma detection with deep learning,'' {\em bioRxiv
  preprint bioRxiv:381855}, 2018.

\bibitem{bi2017ISBI}
L.~Bi, J.~Kim, E.~Ahn, D.~Feng, and M.~Fulham, ``Semi-automatic skin lesion
  segmentation via fully convolutional networks,'' in {\em Biomedical Imaging
  (ISBI 2017), 2017 IEEE 14th International Symposium on}, pp.~561--564, IEEE,
  2017.

\bibitem{van2015transfer}
A.~Van~Opbroek, M.~A. Ikram, M.~W. Vernooij, and M.~De~Bruijne, ``Transfer
  learning improves supervised image segmentation across imaging protocols,''
  {\em IEEE transactions on medical imaging}, vol.~34, no.~5, pp.~1018--1030,
  2015.

\bibitem{weiss2016survey}
K.~Weiss, T.~M. Khoshgoftaar, and D.~Wang, ``A survey of transfer learning,''
  {\em Journal of Big Data}, vol.~3, no.~1, p.~9, 2016.

\bibitem{yu2015multi}
F.~Yu and V.~Koltun, ``Multi-scale context aggregation by dilated
  convolutions,'' {\em arXiv preprint arXiv:1511.07122}, 2015.

\bibitem{deng2009imagenet}
J.~Deng, W.~Dong, R.~Socher, L.-J. Li, K.~Li, and L.~Fei-Fei, ``Imagenet: A
  large-scale hierarchical image database,'' in {\em Computer Vision and
  Pattern Recognition, 2009. CVPR 2009. IEEE Conference on}, pp.~248--255,
  Ieee, 2009.

\bibitem{HUANG2017DENSELY}
G.~Huang, Z.~Liu, L.~Van Der~Maaten, and K.~Q. Weinberger, ``Densely connected
  convolutional networks.,'' in {\em CVPR}, 2017.

\bibitem{chollet2017xception}
F.~Chollet, ``Xception: Deep learning with depthwise separable convolutions,''
  {\em arXiv preprint}, pp.~1610--02357, 2017.

\bibitem{szegedy2017inception}
C.~Szegedy, S.~Ioffe, V.~Vanhoucke, and A.~A. Alemi, ``Inception-v4,
  inception-resnet and the impact of residual connections on learning.,'' in
  {\em AAAI}, vol.~4, p.~12, 2017.

\bibitem{chen2018encoder}
L.-C. Chen, Y.~Zhu, G.~Papandreou, F.~Schroff, and H.~Adam, ``Encoder-decoder
  with atrous separable convolution for semantic image segmentation,'' {\em
  arXiv preprint arXiv:1802.02611}, 2018.

\bibitem{Everingham10}
M.~Everingham, L.~Van~Gool, C.~K.~I. Williams, J.~Winn, and A.~Zisserman, ``The
  pascal visual object classes (voc) challenge,'' {\em International Journal of
  Computer Vision}, vol.~88, pp.~303--338, June 2010.

\bibitem{mirzaalian2014hair}
H.~Mirzaalian, T.~K. Lee, and G.~Hamarneh, ``Hair enhancement in dermoscopic
  images using dual-channel quaternion tubularness filters and mrf-based
  multilabel optimization,'' {\em IEEE Transactions on Image Processing},
  vol.~23, no.~12, pp.~5486--5496, 2014.

\bibitem{glorot2010understanding}
X.~Glorot and Y.~Bengio, ``Understanding the difficulty of training deep
  feedforward neural networks,'' in {\em Proceedings of the thirteenth
  international conference on artificial intelligence and statistics},
  pp.~249--256, 2010.

\bibitem{reddi2018convergence}
S.~J. Reddi, S.~Kale, and S.~Kumar, ``On the convergence of adam and beyond,''
  2018.

\bibitem{codella2017deep}
N.~C. Codella, Q.-B. Nguyen, S.~Pankanti, D.~Gutman, B.~Helba, A.~Halpern, and
  J.~R. Smith, ``Deep learning ensembles for melanoma recognition in dermoscopy
  images,'' {\em IBM Journal of Research and Development}, vol.~61, no.~4,
  pp.~5--1, 2017.

\bibitem{chollet2015keras}
F.~Chollet {\em et~al.}, ``Keras.'' \url{https://keras.io}, 2015.

\bibitem{hajian2013ROC}
K.~Hajian-Tilaki, ``Receiver operating characteristic (roc) curve analysis for
  medical diagnostic test evaluation,'' {\em Caspian journal of internal
  medicine}, vol.~4, no.~2, p.~627, 2013.

\end{thebibliography}

\end{document}